\documentclass[lettersize,journal]{IEEEtran}
\usepackage{amsmath,amssymb, amsfonts}
\usepackage{algorithmic}
\usepackage{algorithm}
\usepackage{array}
\usepackage[caption=false,font=normalsize,labelfont=sf,textfont=sf]{subfig}
\usepackage{textcomp}
\usepackage{stfloats}
\usepackage{url}
\usepackage{verbatim}
\usepackage{graphicx}
\usepackage{cite}
\usepackage{pgfplots}
\pgfplotsset{compat = newest}
\usepgfplotslibrary{groupplots}
\usepgfplotslibrary{colorbrewer}
\usepackage{tikz}
\usetikzlibrary{matrix, positioning,external, fit, backgrounds, patterns, shapes.geometric, arrows.meta, positioning, calc, shadows}
\usepackage{standalone}
\usepackage{xcolor}
\usepackage{tabularray}
\usepackage{multirow}
\usepackage{todonotes}
\usepackage{float}
\usepackage{booktabs}   % professional horizontal rules
\usepackage{tabularx}   % automatic column widths
\usepackage{siunitx}    % aligned numbers
\usepackage{multirow}
\usepackage{standalone}
\usepackage{xcolor}

\definecolor{darkgray176}{RGB}{176,176,176}
\definecolor{lightgray204}{RGB}{204,204,204}
\definecolor{goldenrod1911910}{RGB}{191,191,0}
\definecolor{green01270}{RGB}{0,127,0}
\definecolor{darkorange25512714}{RGB}{255,127,14}
\definecolor{forestgreen4416044}{RGB}{44,160,44}
\definecolor{steelblue31119180}{RGB}{31,119,180}

\definecolor{darkWhite}{rgb}{0.96,0.96,0.96}
\definecolor{bluekeywords}{rgb}{0.13,0.13,1}
\definecolor{greencomments}{rgb}{0,0.5,0}
\definecolor{redstrings}{rgb}{0.9,0,0}

\definecolor{Comment}{RGB}{97,161,176}

\definecolor{btfGreen}{RGB}{51,160,44}
\definecolor{btfRed}{RGB}{190,60,90}

\definecolor{bleuUni}{RGB}{0, 157, 224}
\definecolor{marronUni}{RGB}{68, 58, 49}

\definecolor{bluecite}{HTML}{009DE0}

\definecolor{Paired-1}{RGB}{31,120,180}
\definecolor{Paired-2}{RGB}{166,206,227}
\definecolor{Paired-3}{RGB}{51,160,44}
\definecolor{Paired-4}{RGB}{178,223,138}
\definecolor{Paired-5}{RGB}{227,26,28}
\definecolor{Paired-6}{RGB}{251,154,153}
\definecolor{Paired-7}{RGB}{255,127,0}
\definecolor{Paired-8}{RGB}{253,191,111}
\definecolor{Paired-9}{RGB}{106,61,154}
\definecolor{Paired-10}{RGB}{202,178,214}
\definecolor{Paired-11}{RGB}{177,89,40}
\definecolor{Paired-12}{RGB}{255,255,153}
\definecolor{Accent-1}{RGB}{127,201,127}
\definecolor{Accent-2}{RGB}{190,174,212}
\definecolor{Accent-3}{RGB}{253,192,134}
\definecolor{Accent-4}{RGB}{255,255,153}
\definecolor{Accent-5}{RGB}{56,108,176}
\definecolor{Accent-6}{RGB}{240,2,127}
\definecolor{Accent-7}{RGB}{191,91,23}
\definecolor{Accent-8}{RGB}{102,102,102}
\definecolor{Spectral-1}{RGB}{158,1,66}
\definecolor{Spectral-2}{RGB}{213,62,79}
\definecolor{Spectral-3}{RGB}{244,109,67}
\definecolor{Spectral-4}{RGB}{253,174,97}
\definecolor{Spectral-5}{RGB}{254,224,139}
\definecolor{Spectral-6}{RGB}{255,255,191}
\definecolor{Spectral-7}{RGB}{230,245,152}
\definecolor{Spectral-8}{RGB}{171,221,164}
\definecolor{Spectral-9}{RGB}{102,194,165}
\definecolor{Spectral-10}{RGB}{50,136,189}
\definecolor{Spectral-11}{RGB}{94,79,162}
\definecolor{Set1-1}{RGB}{228,26,28}
\definecolor{Set1-2}{RGB}{55,126,184}
\definecolor{Set1-3}{RGB}{77,175,74}
\definecolor{Set1-4}{RGB}{152,78,163}
\definecolor{Set1-5}{RGB}{255,127,0}
\definecolor{Set1-6}{RGB}{255,255,51}
\definecolor{Set1-7}{RGB}{166,86,40}
\definecolor{Set1-8}{RGB}{247,129,191}
\definecolor{Set1-9}{RGB}{153,153,153}
\definecolor{Set1-10}{RGB}{0,0,0}
\definecolor{Set2-1}{RGB}{102,194,165}
\definecolor{Set2-2}{RGB}{252,141,98}
\definecolor{Set2-3}{RGB}{141,160,203}
\definecolor{Set2-4}{RGB}{231,138,195}
\definecolor{Set2-5}{RGB}{166,216,84}
\definecolor{Set2-6}{RGB}{255,217,47}
\definecolor{Set2-7}{RGB}{229,196,148}
\definecolor{Set2-8}{RGB}{179,179,179}
\definecolor{Dark2-1}{RGB}{27,158,119}
\definecolor{Dark2-2}{RGB}{217,95,2}
\definecolor{Dark2-3}{RGB}{117,112,179}
\definecolor{Dark2-4}{RGB}{231,41,138}
\definecolor{Dark2-5}{RGB}{102,166,30}
\definecolor{Dark2-6}{RGB}{230,171,2}
\definecolor{Dark2-7}{RGB}{166,118,29}
\definecolor{Dark2-8}{RGB}{102,102,102}
\definecolor{Reds-1}{RGB}{255,245,240}
\definecolor{Reds-2}{RGB}{254,224,210}
\definecolor{Reds-3}{RGB}{252,187,161}
\definecolor{Reds-4}{RGB}{252,146,114}
\definecolor{Reds-5}{RGB}{251,106,74}
\definecolor{Reds-6}{RGB}{239,59,44}
\definecolor{Reds-7}{RGB}{203,24,29}
\definecolor{Reds-8}{RGB}{165,15,21}
\definecolor{Reds-9}{RGB}{103,0,13}
\definecolor{Greens-1}{RGB}{247,252,245}
\definecolor{Greens-2}{RGB}{229,245,224}
\definecolor{Greens-3}{RGB}{199,233,192}
\definecolor{Greens-4}{RGB}{161,217,155}
\definecolor{Greens-5}{RGB}{116,196,118}
\definecolor{Greens-6}{RGB}{65,171,93}
\definecolor{Greens-7}{RGB}{35,139,69}
\definecolor{Greens-8}{RGB}{0,109,44}
\definecolor{Greens-9}{RGB}{0,68,27}
\definecolor{Blues-1}{RGB}{247,251,255}
\definecolor{Blues-2}{RGB}{222,235,247}
\definecolor{Blues-3}{RGB}{198,219,239}
\definecolor{Blues-4}{RGB}{158,202,225}
\definecolor{Blues-5}{RGB}{107,174,214}
\definecolor{Blues-6}{RGB}{66,146,198}
\definecolor{Blues-7}{RGB}{33,113,181}
\definecolor{Blues-8}{RGB}{8,81,156}
\definecolor{Blues-9}{RGB}{8,48,107}

\begin{document}

\title{MONET: Modeling and Optimization of neural NEtwork Training from Edge to Data Centers}

\author{Jérémy~Morlier,
        Arne~Symons,
        Robin~Geens,
        Stef~Cuyckens,
        Marian~Verhelst,
        Vincent~Gripon,
        and Mathieu~Léonardon%
\thanks{J. Morlier, M. Léonardon, and V. Gripon are with IMT Atlantique, Brest, France.}%
\thanks{R. Geens, S. Cuyckens, M. Verhelst, and A. Symons are with MICAS, KU Leuven, Leuven, Belgium.}%
}
% Jérémy Morlier
% Robin Geens (MICAS, KU Leuven)
% Stef Cuyckens (MICAS, KU Leuven)
% Marian Verhelst (MICAS, KU Leuven)
% Arne Symons (MICAS, KU Leuven)
% Mathieu Léonardon
% Vincent Gripon

\markboth{Journal of \LaTeX\ Class Files,~Vol.~14, No.~8, August~2021}%
{Shell \MakeLowercase{\textit{et al.}}: A Sample Article Using IEEEtran.cls for IEEE Journals}

\IEEEpubid{0000--0000/00\$00.00~\copyright~2021 IEEE}

\maketitle

\begin{abstract}
While hardware-software co-design has significantly improved the efficiency of neural network inference, modeling the training phase remains a critical yet underexplored challenge. Training workloads impose distinct constraints, particularly regarding memory footprint and backpropagation complexity, which existing inference-focused tools fail to capture. 
This paper introduces MONET, a framework designed to model the training of neural networks on heterogeneous dataflow accelerators. MONET builds upon Stream, an experimentally verified framework that that models the inference of neural networks on heterogeneous dataflow accelerators with layer fusion. Using MONET, we explore the design space of ResNet-18 and a small GPT-2, demonstrating the framework's capability to model training workflows and find better hardware architectures.
We then further examine problems that become more complex in neural network training due to the larger design space, such as determining the best layer-fusion configuration. Additionally, we use our framework to find interesting trade-offs in activation checkpointing, with the help of a genetic algorithm. Our findings highlight the importance of a holistic approach to hardware-software co-design for scalable and efficient deep learning deployment.
\end{abstract}


\section{Introduction}

Deep Neural Networks (DNNs) have revolutionized fields ranging from computer vision to natural language processing. Their success hinges on the ability to learn complex patterns from large volumes of data. 
% Training involves iteratively adjusting millions or even billions of parameters to minimize a loss function, which demands substantial computational resources and energy.

Although optimizing the resources needed for inference has received significant attention in recent years~\cite{groq2022lpu, Cerebras2023WSE}, training remains underexplored yet a critical and resource-intensive phase in the life cycle of deep neural networks. Training cost is indeed substantial across a wide spectrum of deployment settings: from edge environments, such as personalized healthcare, where data privacy and real-time adaptation are essential, to large-scale datacenters training massive models like Large Language Models (LLMs) on clusters of GPUs or TPUs~\cite{google2020tpuv2}. The rapidly increasing cost of training LLMs~\cite{cottier2024rising}, coupled with the substantial environmental impact of the supporting infrastructure and energy consumption~\cite{luccioni2023estimating}, has intensified the need to improve training efficiency, a key motivation for our work. This problem is inherently multi-dimensional, requiring careful consideration of factors such as neural network architecture, training hyperparameters, hardware platform characteristics, and deployment strategies like data/model parallelism, scheduling, and layer fusion.

\begin{figure}[t!]
\begin{center}
\def\colormap{viridis}
\def\colorData{totalCompln}
\def\colorName{\shortstack{Total\\Computational\\Capacity}}

\begin{tikzpicture}
\begin{groupplot}[
    group style={group size=1 by 2, vertical sep=1.5cm},
    width=0.85\linewidth,
    height=0.5\linewidth,
    tick align=outside,
    tick pos=left,
    x grid style={darkgray176!60},
    y grid style={darkgray176!60},
    x grid style={},
    y grid style={},
    xmajorgrids,
    grid=major,
    grid style={dashed},
    xtick style={color=black},
    scaled x ticks=base 10:-9,
    xtick scale label code/.code={},
    xlabel={Energy ($\times 10^{9}$ pJ)},
    ytick style={color=black},
    ylabel near ticks,
    ylabel={Latency (Cycles)},
    scaled y ticks=base 10:-7,
    ytick scale label code/.code={},
    ylabel={Latency ($\times 10^{7}$ Cycles)},
    colormap/\colormap,
    point meta=explicit,
    legend cell align={left},
    legend style={
      fill opacity=0.8,
      draw=black,
      draw opacity=1,
      text opacity=1,
      at={(0.01,0.98)},
      anchor=north west,
      draw=lightgray204
    },
]

% --- Top: Inference (NO colorbar) ---
\nextgroupplot[
    title=\it{Inference},
    colorbar,
    xlabel={},
    % xmin=2e9, xmax=4e9,
    % ymin=0, ymax=0.8e8,
    ytick={ 6.62, 7.01, 7.39, 7.77, 8.15 },
    yticklabels={ 0, 1, 2, 6, 14 },
    xtick={ 9.38, 9.45, 9.53, 9.61, 9.69 },
    xticklabels={ 2, 3, 3, 4, 5 },
    colorbar style={
        ytick style={draw=none},
        title=\shortstack{Total\\Computational\\Capacity},
        title style={align=center},
        height=0.7\linewidth + 1cm,
        scaled y ticks=false, % <- do not apply scaled y ticks to colorbar
        ytick scale label code/.code={}, % <- no "×10^k" scale label on the colorbar
        ytick= {1.20411998265592, 1.901604224927816, 2.599088467199712, 3.296572709471608, 3.994056951743504, 4.6915411940154},
        yticklabels={16, 80, 397, 1980, 9864, 49152},
    },
]

\addplot [thick, only marks, mark=*, mark size=1.5, scatter]
table [y=f_log_latency, x=f_log_energy, meta=\colorData, col sep=comma]
{figures/exploreHardware/result.csv};

% --- Bottom: Training (the ONLY one with colorbar) ---
\nextgroupplot[
    title=\it{Training},
    ytick={ 7.40, 7.71, 8.02, 8.33, 8.64 },
    yticklabels={ 3, 5, 11, 21, 43 },
    xtick={ 10.42, 10.55, 10.67, 10.79, 10.91 },
    xticklabels={ 26, 35, 47, 62, 82 },
]
\addplot [thick, only marks, mark=*, mark size=1.5, scatter]
table [y=fb_log_latency, x=fb_log_energy, meta=\colorData, col sep=comma]
{figures/exploreHardware/result.csv};

\end{groupplot}
\end{tikzpicture}
\end{center}
\vspace{-0.5cm}
\caption{Energy–latency trade-offs of a ResNet-18 across a diverse range of Edge TPU architectural configurations. Results are shown for inference (top) and training (bottom). Each point represents a distinct hardware configuration, with energy consumption (pJ) plotted against execution latency (cycles) and color-coded by total computational capacity. This highlights the energy/latency distribution difference between inference and training and the need to model hardware specifically for training purposes.}
\label{fig:main}
\end{figure}
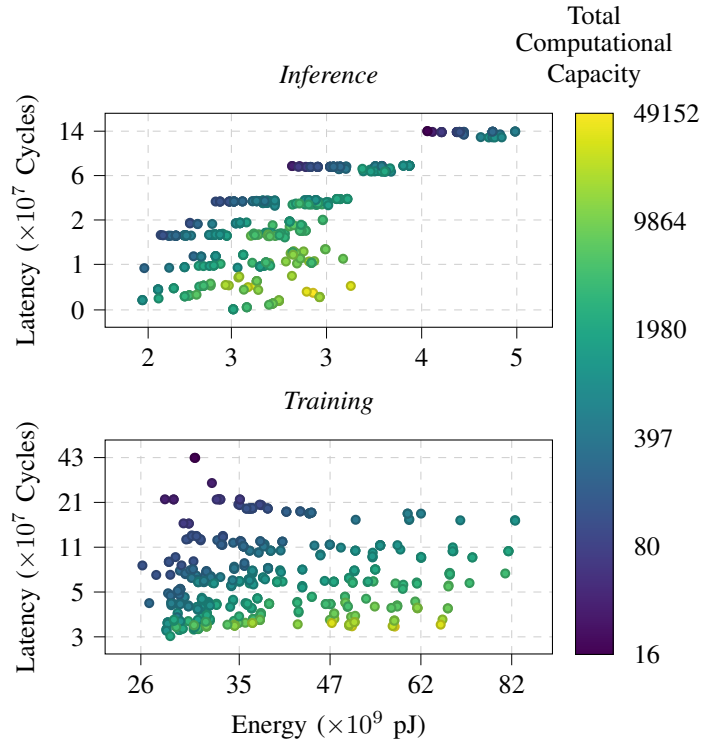
Among the important dimensions, architectural innovations such as ResNet~\cite{he2016deep} to the more recent Transformer~\cite{vaswani2017attention} and State-Space Model~\cite{guefficiently} have allowed for fast and energy-efficient training of neural networks. Additionally, hyperparameter optimization~\cite{franceschi2024hyperparameter,li2018hyperband} such as tuning learning rates, batch sizes and optimizer~\cite{liu2025muon, kingma2014adam} plays an important role in accelerating convergence and reducing the computational cost.
\IEEEpubidadjcol

At the heart of efficient training on custom hardware lies the accelerator architecture. The evolution of hardware has driven the increasing demands of neural network training. GPUs (\emph{e.g.}, NVIDIA H100~\cite{nvidia_h100_whitepaper}, AMD MI300A~\cite{amd2024mi300A}) and TPUs (\emph{e.g.}, Google TPU v2~\cite{google2020tpuv2}, TPU v4~\cite{google2023TPUv4}) are specifically designed to process deep learning workloads, offering massive parallelism and optimized dataflows for matrix operations. More recently, architectures such as the Cerebras WaferScale Engine (WSE)~\cite{Cerebras2023WSE} and Groq LPU~\cite{groq2022lpu} have pushed the boundaries further by eliminating traditional memory bottlenecks and enabling unprecedented levels of on-chip communication and computation, making them particularly interesting for Large Language Model (LLM) decoding.

Furthermore, efficient deployment of deep learning training workloads goes beyond selecting the right neural network architecture or hardware platform. It requires carefully orchestrating computations between compute cores, and time. Key aspects of this deployment dimension include parallelism strategies (data, pipeline, and tensor parallelism), scheduling of computations to optimize hardware utilization, and memory management techniques such as activation checkpointing. Together, these strategies form a crucial layer in the overall optimization of neural network training.

% As a result, the search space of hardware accelerators for neural network training is vast, highly irregular, and difficult to explore efficiently. Each of the dimensions introduces trade-offs in terms of energy consumption, throughput, and latency. Before optimization can even be attempted, it is crucial to develop fast and accurate modeling tools capable of estimating these key performance metrics.

% SOTA dataflow modelisations
In short, designing hardware accelerators and the associated deployment strategies for neural network training inherently involves complex trade-offs among energy consumption, throughput, and latency. Figure~\ref{fig:main} highlights that hardware configurations exhibit distinct energy–latency distributions depending on whether they are evaluated using inference or training workloads, demonstrating that conclusions drawn from inference-only analysis do not necessarily transfer to training scenarios and reinforcing the need for training-aware evaluation. Effectively navigating these trade-offs requires the ability to quickly and accurately model the corresponding performance metrics, enabling more informed hardware design and configuration decisions.
In this work, we tackle the central question of how to model the most effective hardware configurations and deployment configurations for efficiently training neural networks across diverse scenarios, aiming to reduce both financial costs and energy requirements. To this aim, we make the following contributions:

\begin{itemize}
    \item We propose MONET, a modeling tool for training-optimized hardware by extending Stream~\cite{symons2024stream}, a modeling and scheduling framework for estimating the performance of DNNs on heterogeneous dataflow accelerators, to support training workloads.
    
    \item We illustrate the irregularity of the training–hardware–deployment design space through two representative examples: ResNet-18 on an image classification task, and a small GPT-2 on an NLP task.
    
    \item We propose a constraint optimization approach to explore the fused-layer deployment search space and discover improved configurations.
    
    \item We leverage our extended framework to study the activation checkpointing problem and propose a genetic algorithm to efficiently solve it.
\end{itemize}

The remainder of this paper is organized as follows.
Section~\ref{sec:background} briefly introduces the main concepts needed for this work, including neural network training, memory overheads, heterogeneous dataflow accelerators, and deployment strategies, and reviews related modeling frameworks.
Sections~\ref{lb:framework} and~\ref{lb:SearchSpace} introduce our extended modeling framework and characterize the joint training--hardware--deployment design space. Through two case studies (ResNet-18 and GPT-2), we illustrate the complexity and workload-dependent nature of training-aware optimization.
Section~\ref{pag:lfs} presents our constraint-based layer-fusion algorithm, while Section~\ref{pag:AC} addresses activation checkpointing and proposes a genetic algorithm to handle its inherent non-linearity.
% Section~\ref{sec:synthesis} combines these techniques and evaluates their joint impact on latency, energy, and memory. 
Finally, Section~\ref{sec:conclusion} concludes the paper and outlines future work.

\section{Background}\label{sec:background}

In this section, we present three key dimensions of the design space for the training of neural networks from three different angles: the training process itself, the hardware architecture and the deployment strategies. We give for each of these angles a brief overview of the key concepts and most importantly how these choices can impact the metrics we want to model: latency, throughput and energy consumption.
Finally, we discuss existing DNN modeling tools and introduce Stream, which we extend to support training workloads across various hardware architectures.
\begin{figure}[b]
    \centering
    \includegraphics[width=\linewidth]{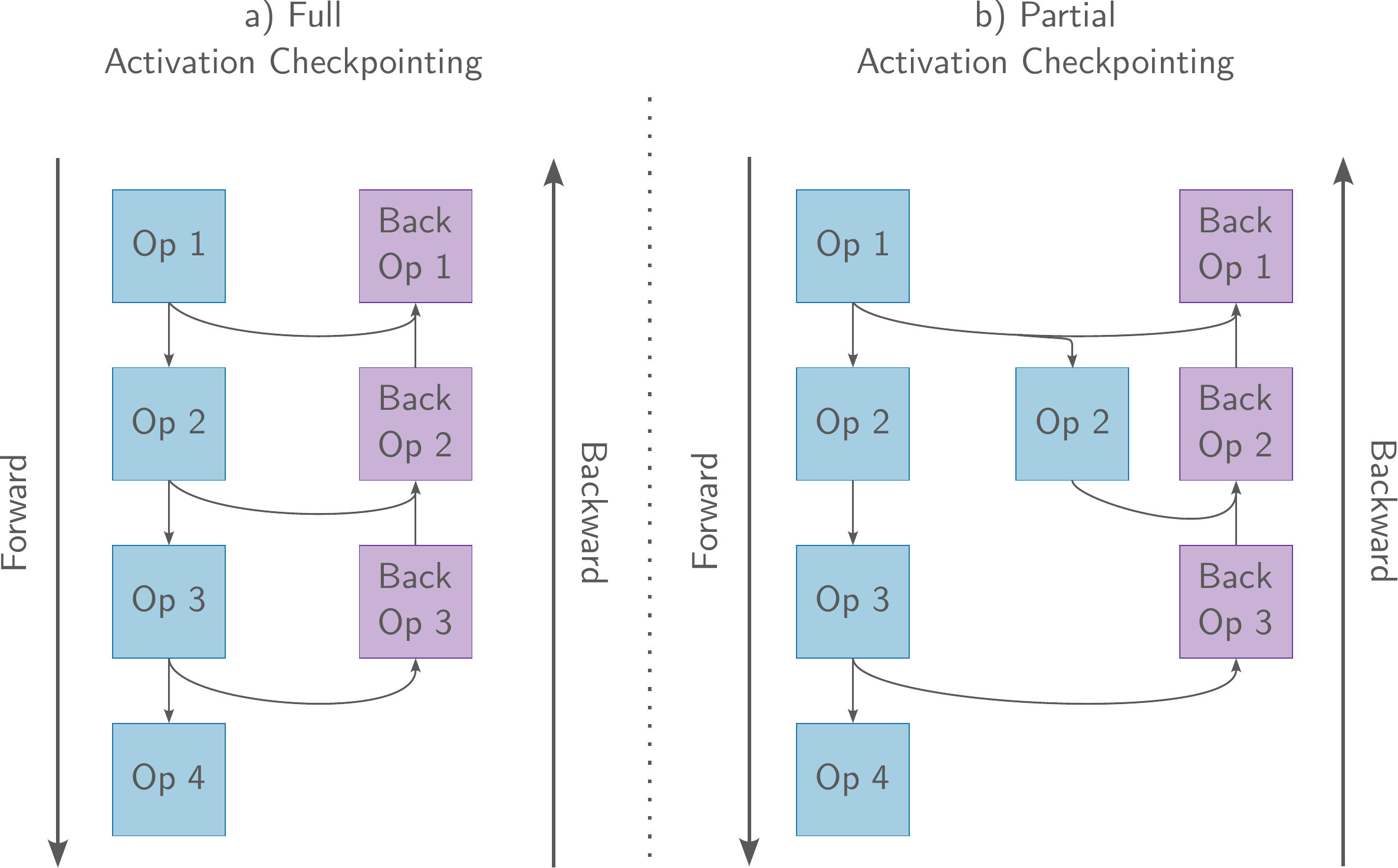}
    \caption{(a) All activations (edges) are saved during the forward pass and reused for the backward pass. (b) Some activations are discarded and recomputed during the backward pass, reducing the total memory cost.}
    \label{fig:activationCheckpointing}
\end{figure}

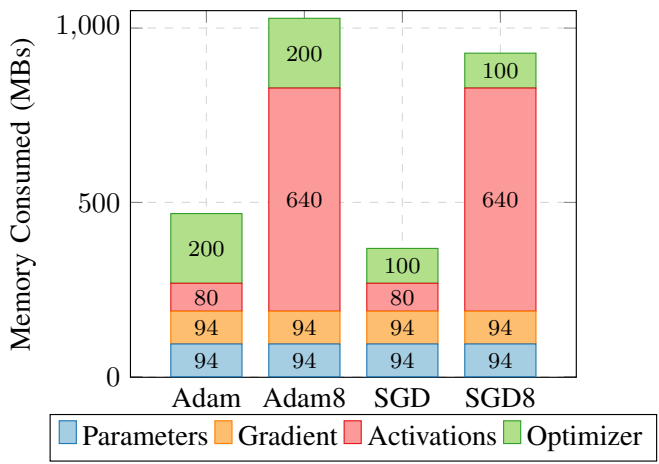
\begin{figure}[t]
    \centering
    \resizebox{\linewidth}{!}{    
    \begin{tikzpicture}
    \begin{axis}[
        ybar stacked,
        bar width=25pt,
        nodes near coords,
        width=0.8\columnwidth,
        nodes near coords style={text=black, font=\footnotesize},
        xmin=0,
        xmax=14,
        enlarge y limits={lower=0},
        enlarge x limits=0.15,
        ymax=1050,
        legend style={at={(0.5,-0.10)},
        anchor=north,legend columns=-1},
        ylabel={Memory Consumed (MBs)},
        grid=major,
        grid style={dashed, gray!30},
        xtick={1, 5, 9, 13},
        xticklabels={Adam, Adam8, SGD, SGD8},
        % symbolic x coords={Adam, SGD},
        ]
    \addplot+[ybar, fill=Paired-2, draw=Paired-1] plot coordinates {(1, 94) (5, 94) (9, 94) (13, 94)};
    \addplot+[ybar, fill=Paired-8, draw=Paired-7] plot coordinates {(1, 94) (5, 94) (9, 94) (13, 94)};
    \addplot+[ybar, fill=Paired-6, draw=Paired-5] plot coordinates {(1, 80) (5, 640) (9, 80) (13, 640)};
     \addplot+[ybar, fill=Paired-4, draw=Paired-3] plot coordinates {(1, 200) (5, 200) (9, 100) (13, 100)};
    \legend{Parameters, Gradient,  Activations, Optimizer}
    \end{axis}
    \end{tikzpicture}}
    \caption{Peak memory consumption (GBs) breakdown for a Resnet-50 measured on an RTX3090 with an image of size 224 by 224, with two different batch sizes, (1 and 8).}
    \label{fig:membreakdown}
\end{figure}

\subsection{Training}
Training a neural network is a computationally intensive process comprising three fundamental steps: forward pass, backward pass, and weight update. 

In this paper, we model a neural network as a directed graph \(\mathcal{G} = (\mathcal{V}, \mathcal{E})\), where \(\mathcal{V}\) represents the set of \(N\) nodes corresponding to the network's operators (or layers), and \(\mathcal{E}\) denotes the set of edges representing the tensors exchanged as inputs and outputs between these operators.

Given an input vector \(\mathbf{x}\), the forward pass computes the output of each layer sequentially using a parameterized function~\(f\). For a layer \(l \in \mathcal{V}\), the output \(\mathbf{a}^{(l)}\) is determined as:
\begin{equation}\label{eq:forward}
    \mathbf{a}^{(l)} = f^{(l)}(\mathbf{a}^{(l-1)}, \theta^{(l)})
\end{equation}
where:
\begin{itemize}
    \item \(\mathbf{a}^{(0)} = \mathbf{x}\) is the input to the network,
    \item \(\theta^{(l)}\) are the parameters (weights and biases) of layer \(l\),
    \item \(f^{(l)}\) is the transformation function for layer \(l\), which may include linear transformations and activation functions.
\end{itemize}

The backward pass computes the gradients of the loss function \(\mathcal{L}\) with respect to the parameters \(\theta^{(l)}\) using backpropagation. For a layer \(l\), the gradient of the loss with respect to the input \(\mathbf{a}^{(l-1)}\) is given by:
\begin{equation}\label{eq:inputGradient}
    \frac{\partial \mathcal{L}}{\partial \mathbf{a}^{(l-1)}} = \frac{\partial \mathcal{L}}{\partial \mathbf{a}^{(l)}} \cdot \frac{\partial f^{(l)}(\mathbf{a}^{(l-1)}; \theta^{(l)})}{\partial \mathbf{a}^{(l-1)}}
\end{equation}
The gradient of the loss with respect to the parameters \(\theta^{(l)}\) is computed as:
\begin{equation}\label{eq:weightGradient}
    \frac{\partial \mathcal{L}}{\partial \theta^{(l)}} = \frac{\partial \mathcal{L}}{\partial \mathbf{a}^{(l)}} \cdot \frac{\partial f^{(l)}(\mathbf{a}^{(l-1)}, \theta^{(l)})}{\partial \theta^{(l)}}
\end{equation}
The combined forward and backward passes are illustrated in Figure~\ref{fig:activationCheckpointing}(a).

The optimizer updates the parameters \(\theta^{(l)}\) using the computed gradients. For example, in Stochastic Gradient Descent (SGD) with learning rate \(\eta\) and momentum coefficient \(\mu\), the update rule for the momentum term \(v^{(l)}\) and parameters \(\theta^{(l)}\) is:
\begin{align}\label{eq:optimizer}
    v^{(l)} &\leftarrow \mu v^{(l)} - \eta \cdot \frac{\partial \mathcal{L}}{\partial \theta^{(l)}} \\
    \theta^{(l)} &\leftarrow \theta^{(l)} + v^{(l)}
\end{align}

Furthermore, using methods such as momentum in optimizers introduces additional parameters, referred to as optimizer states (\emph{e.g.}, the momentum term), for each parameter. This increases the memory cost of training neural networks. As shown in Figure~\ref{fig:membreakdown}, optimizers like ADAM~\cite{kingma2014adam} use even more optimizer states, further escalating memory usage beyond that of the model parameters.

Several studies have addressed this memory overhead. For instance, Galore~\cite{zhao2024galore} reduces the memory footprint by applying the optimizer to a low-rank approximation of the weight gradients rather than the gradients themselves. This approach effectively decreases the number of optimizer states required per weight, thereby lowering the overall memory cost.

Another significant contributor to memory cost is the storage of activations. Unlike inference, where activations are typically used in subsequent layers or residual connections, training requires most activations generated during the forward pass to be retained for gradient computation, as shown in Equation~\eqref{eq:weightGradient}. This results in a substantially higher memory demand compared to inference, particularly as the batch size increases, as illustrated in Figure~\ref{fig:membreakdown}. Leveraging the specific properties of gradient computation, activations can be stored in a compressed format tailored to the operation being backpropagated as studied in Gist~\cite{Jain2018Gist}. For example, the backpropagation of ReLU only requires the sign of its activation output to propagate the gradient, reducing the memory required for storing activations.

%TODO: text too small ?\as{Text is too small.}
\begin{figure}[b!]
    \centering
    \includegraphics[width=0.6\linewidth]{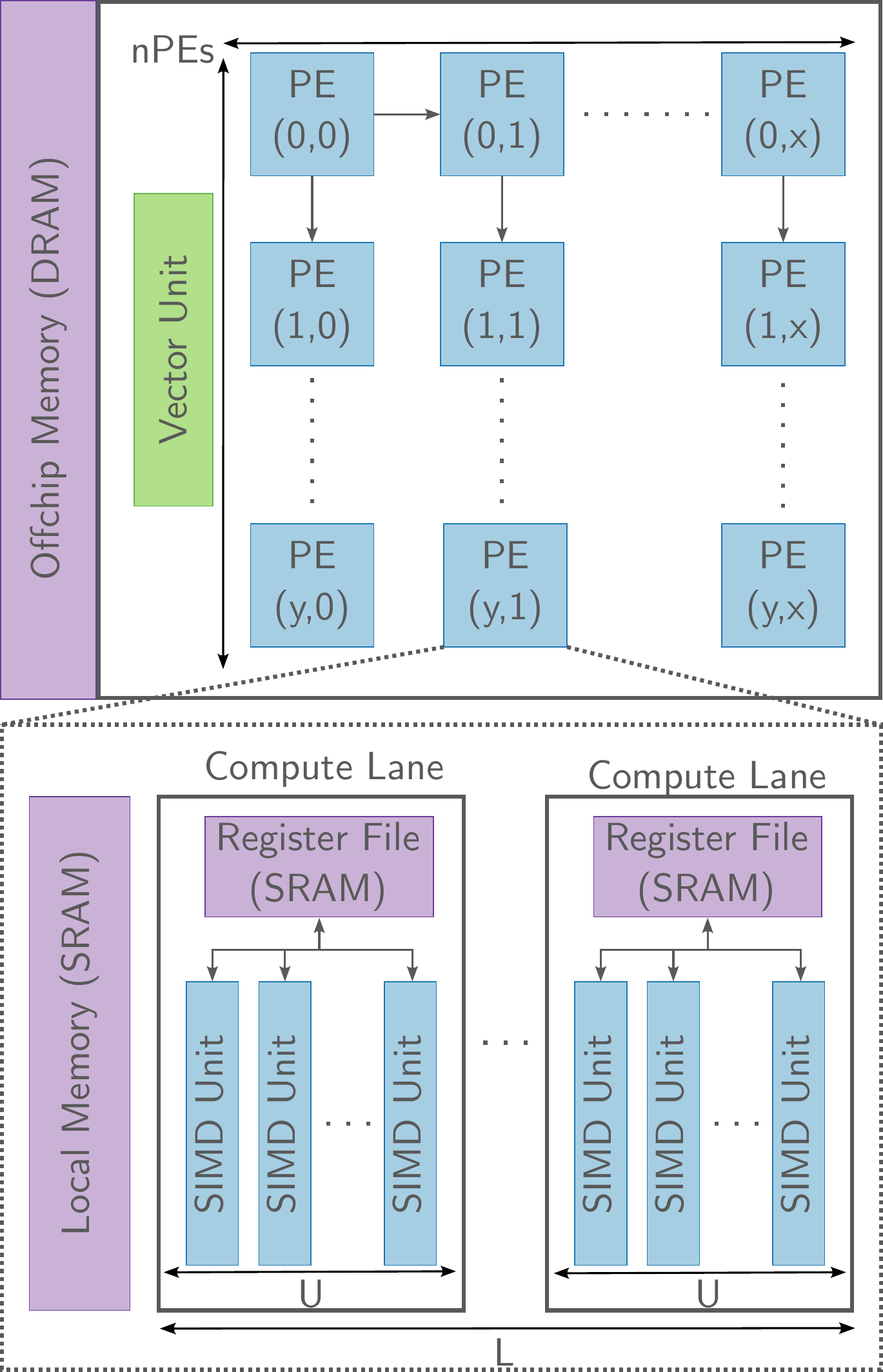}
    \caption{EdgeTPU architecture: A set of $n_{PEs}$ Processing Elements (PE) is arranged in a 2D array, each capable of communicating with its neighbours and a common bus link all of them to an off-chip memory.  Each PE is composed of a memory and a weight stationary accelerator composed of $U$ SIMD Units and $L$ Compute Lanes. Adapted from~\cite{Zhou2022CoDesignTPU}.}
    \label{fig:edgetpuarchitecture}
\end{figure}

\begin{figure*}[t]
    \centering
    \includegraphics[width=\linewidth]{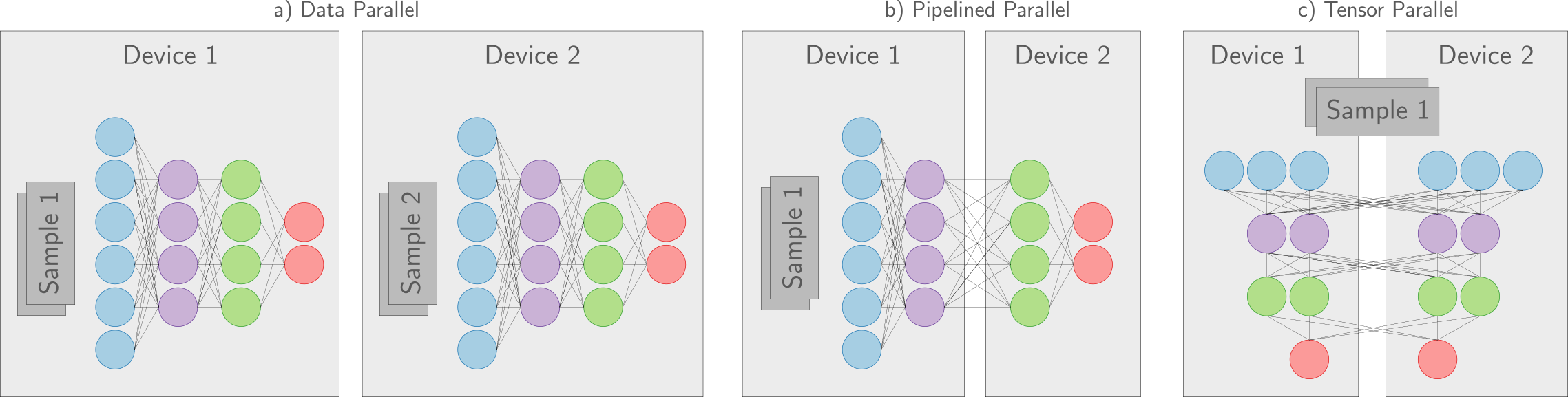}
    \caption{Three strategies for spatially parallelizing deep learning workloads across devices~\cite{Cerebras2023WSE}:
        (a) Data Parallel: Distributes the batch dimension across multiple devices, with each device processing a distinct subset of samples.
        (b) Pipelined Parallel: Partitions the model into stages, with each device responsible for executing a specific stage of the computation.
        (c) Tensor Parallel: Splits individual layers or operations across devices, enabling parallel processing of a single sample.
    % \as{Text is too small.}
    }
    \label{fig:parallelism}
\end{figure*}

% \subsubsection{Activation Checkpointing}
The memory overhead associated with storing all activations for backpropagation can be mitigated through techniques such as activation checkpointing.
Activation checkpointing reduces the memory footprint of activations during the training process by selectively storing only a subset of activations and recomputing others as needed.
By strategically checkpointing a subset of activations and recomputing the remainder, this method significantly reduces memory consumption.
Figure~\ref{fig:activationCheckpointing} illustrates this process: (a) depicts the conventional approach of storing all activations, while (b) demonstrates checkpointing, where certain activations are discarded and recomputed, thereby reducing memory overhead.

The selection of which activations to checkpoint and which to recompute is a critical aspect of activation checkpointing.
This selection introduces a trade-off between memory savings—achieved by recomputing a larger number of activations—and the associated recomputation cost.
Inductor~\cite{pytorch_activation_checkpointing_2025}, a PyTorch compiler, leverages operator characteristics to discard activations resulting from element-wise operators, as these operators are typically memory-bound and can be efficiently recomputed when fused with preceding operators.
This approach can result in a reduction of both memory cost and runtime.
However, eliminating element-wise operators that require computed activations inherently limits the potential memory savings for neural network training.

To formalize the activation checkpointing optimization problem, frameworks such as Dace-AD~\cite{boudaoud2025dace} and Checkmate~\cite{jain2020checkmate} model it as a mixed integer linear programming (MILP) problem.
In this formulation, each activation—represented as an edge between the forward and backward passes—is associated with two primary costs:
its memory cost \( m_a \) (in bytes), required to store the activation, and its recompute cost \( r_a \) (in floating-point operations, (FLOPs)), required to recompute it during the backward pass.

For a given neural network, let \(\mathcal{A}\) denote the set of activations required for the backward pass.
We introduce a binary decision variable \( x_a \in \{0, 1\} \) for each activation \( a \in \mathcal{A} \), where \( x_a = 1 \) indicates that activation \( a \) is checkpointed (saved), and \( x_a = 0 \) indicates that it is recomputed during the backward pass.

The optimization objective is to minimize the total recompute cost while adhering to a memory budget constraint \( M \):
\begin{equation}\label{eq:MILP}
\begin{aligned}
\min_{\{x_a\}_{a \in \mathcal{A}}} \quad & \sum_{a \in \mathcal{A}} r_a \, (1 - x_a) \\
\text{subject to} \quad & \sum_{a \in \mathcal{A}} m_a \, x_a \le M.
\end{aligned}
\end{equation}

This MILP formulation systematically explores the trade-offs between memory usage and recomputation overhead.
By solving this problem, the approach can be adapted to a wide range of hardware configurations for neural network training.
However, a limitation of this model is its reliance on FLOPs as a proxy for recompute cost, which may not fully account for operator-specific characteristics, as explored by Inductor~\cite{pytorch_mincut_2022}.

\subsection{Hardware}
Designing modern hardware architectures for deep learning is inherently complex due to the need to precisely orchestrate how data is produced, moved, and consumed across compute and memory resources. As architectures scale and diversify, performance and efficiency are increasingly determined not only by peak compute capability, but by the chosen dataflow, \emph{i.e.} the spatial and temporal mapping of operations and data movement. Capturing this complexity while retaining analytical tractability is challenging, as real systems exhibit intricate interactions between memory hierarchies, compute units, and control mechanisms.

In this work, we adopt the abstraction of Heterogeneous Dataflow Accelerators (HDAs)~\cite{kwon2021HDA} to reason about modern architectures. Under this definition, an HDA is modeled as a collection of dataflow accelerators interconnected through buses or point-to-point links, as illustrated in Figure~\ref{fig:edgetpuarchitecture} and Figure~\ref{fig:fusemax}. Each accelerator is specialized for a particular class of workloads through a distinct dataflow and a customized memory hierarchy.

Although this abstraction does not explicitly capture the full complexity of control mechanisms found in architectures such as GPUs or TPUs, it effectively represents their dominant computation and data-movement patterns. Prior work, such as LLMCompass~\cite{LLMCompass}, has demonstrated that dataflow-centric abstractions can faithfully approximate the performance characteristics of GPU-like architectures. Consequently, this modeling approach enables systematic exploration of the joint architecture–mapping design space and provides valuable insight into the fundamental trade-offs underlying modern dataflow-oriented systems.
Two key components define each dataflow accelerator within an HDA:
\begin{itemize}
    \item Memory Hierarchy: A multi-level organization (\emph{e.g.}, register files, on-chip buffers, DRAM) designed to reduce data movement and exploit reuse. Memory levels may be shared or operand-specific (weights, inputs, outputs), with sizes and bandwidths chosen to balance on-chip capacity against off-chip access costs.
    \item Spatial Array of Processing Elements (PEs): A multidimensional array of processing elements that execute operations such as multiply-accumulate (MAC) or any arithmetic operations according to a prescribed dataflow.
\end{itemize}

\subsection{Deployment}
In order to leverage the full potential of a given hardware, a fine-grained orchestration is required to mitigate data-movement bottlenecks and maximize resource utilization. Scheduling, in this context, refers to the organization of computations—both spatially and temporally—to optimize performance, resource utilization, and memory access patterns, as defined by frameworks such as Halide~\cite{halide2016}.

In this paper, we focus on two key aspects of scheduling: parallelism, encompassing both spatial and temporal dimensions, which leverages the intrinsic concurrency of hardware architectures, and layer-fused scheduling~\cite{alwani2016layerfused}, which uses cross-layer data dependencies to minimize memory transfers and reduce the runtime of these layers. Additional techniques, such as tiling, are automatically handled by the framework we employ.
\subsubsection{Parallelism}
To fully exploit the capabilities of modern accelerators, several parallelism strategies can be employed, each with distinct trade-offs. In this context, a device can be defined either as multiple HDAs or as individual accelerators within an HDA.
\begin{itemize}
    \item Data Parallelism (Figure~\ref{fig:parallelism}(a) distributes the batch dimension across multiple devices. This approach minimizes inter-device communication but requires each device to store a full copy of the model parameters, which can limit scalability for large models.
    \item Pipeline Parallelism~\cite{huang2019gpipe} (Figure~\ref{fig:parallelism}(b) partitions the model across devices, reducing the per-device memory footprint. However, it introduces the need to transfer intermediate activations between devices, which can become a bottleneck for latency-sensitive workloads.
    \item Tensor Parallelism~\cite{Krizhevsky2012AlexNet} (Figure~\ref{fig:parallelism}(c) splits individual layers across devices, maximizing parallelism but often at the cost of increased communication overhead due to frequent synchronization.
\end{itemize}
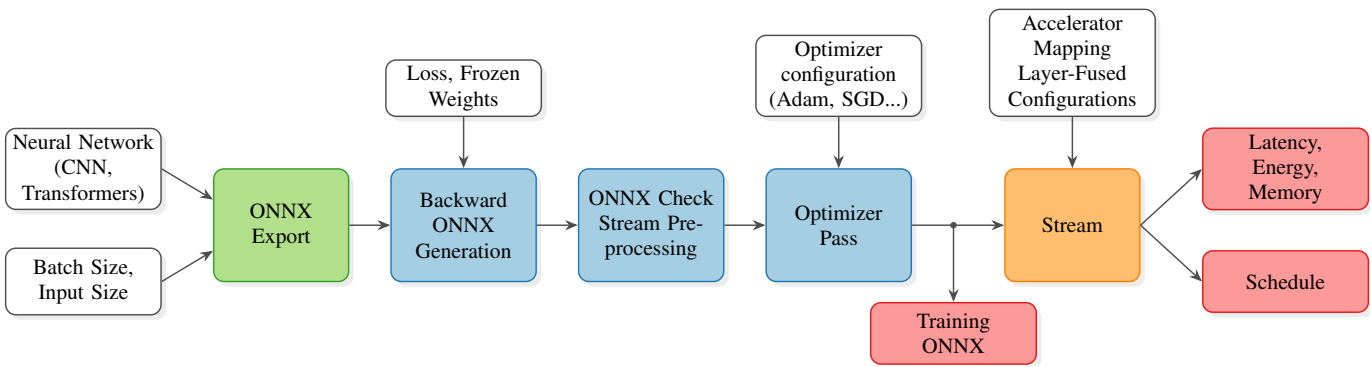
\begin{figure*}[b!]

\centering
\begin{tikzpicture}[
    % Define consistent styles for the diagram
    base/.style={align=center, font=\large, drop shadow={opacity=0.15, shadow xshift=1mm, shadow yshift=-1mm}, thick},
    % Input nodes (Left)
    % Input nodes (Left) - was ellipse -> rectangle
    input/.style={
        base,
        rectangle,
        rounded corners=2mm,
        draw=black!70,
        fill=white,
        text width=2.8cm,
        minimum height=1.3cm
    },
    % Process nodes (Green, Blue, Orange)
    processG/.style={base, rectangle, rounded corners=2mm, draw=Paired-3, fill=Paired-4, text width=2.4cm, minimum height=2.2cm},
    processB/.style={base, rectangle, rounded corners=2mm, draw=Paired-1, fill=Paired-2, text width=2.6cm, minimum height=2.2cm},
    processO/.style={base, rectangle, rounded corners=2mm, draw=Paired-7, fill=Paired-8, text width=2.4cm, minimum height=2.2cm},
    % Top Configuration nodes - was ellipse/circle -> rectangle
    topInputE/.style={
        base,
        rectangle,
        rounded corners=2mm,
        draw=black!70,
        fill=white,
        text width=2.8cm,
        minimum height=1cm
    },
    topInputC/.style={
        base,
        rectangle,
        rounded corners=2mm,
        draw=black!70,
        fill=white,
        text width=3cm,
        minimum height=1.2cm
    },
    % Output nodes (Right & Bottom) - was ellipse -> rectangle
    output/.style={
        base,
        rectangle,
        rounded corners=2mm,
        draw=Paired-5,
        fill=Paired-6,
        text width=3cm,
        minimum height=1.2cm
    },
    % Connecting Arrows
    arrow/.style={-{Stealth[length=2.5mm, width=2mm]}, thick, draw=black!70}
]

    % --- Place main horizontal processing spine ---
    \node[processG] (export) {ONNX\\Export};
    \node[processB, right=0.8cm of export] (backward) {Backward\\ONNX\\Generation};
    \node[processB, right=0.8cm of backward] (check) {ONNX Check\\Stream Pre-processing};
    \node[processB, right=0.8cm of check] (opt) {Optimizer\\Pass};
    \node[processO, right=1.8cm of opt] (stream) {Stream}; % Extra space for the branch

    % --- Place left initial inputs ---
    \node[input, left=1cm of export, yshift=1.1cm] (nn) {Neural Network\\(CNN,\\Transformers)};
    \node[input, left=1cm of export, yshift=-1.1cm] (batch) {Batch Size,\\Input Size};

    % --- Place top configuration inputs ---
    \node[topInputE, above=1cm of backward] (loss) {Loss, Frozen\\Weights};
    \node[topInputC, above=1cm of opt] (opt_cfg) {Optimizer\\configuration\\(Adam, SGD...)};
    \node[topInputC, above=1cm of stream] (accel_cfg) {Accelerator\\Mapping\\Layer-Fused\\Configurations};

    % --- Place bottom/right outputs ---
    % Calculate midpoint for the branching output
    \coordinate (branch_pt) at ($(opt.east)!0.45!(stream.west)$);
    \node[output, below=1.5cm of branch_pt] (train) {Training\\ONNX};
    
    \node[output, right=1.2cm of stream, yshift=1.1cm] (latency) {Latency,\\Energy,\\Memory};
    \node[output, right=1.2cm of stream, yshift=-1.1cm] (schedule) {Schedule};

    % --- Draw Connections ---
    % Left Inputs to ONNX Export
    \draw[arrow] (nn.east) -- ([yshift=0.5cm]export.west);
    \draw[arrow] (batch.east) -- ([yshift=-0.5cm]export.west);

    % Main horizontal spine
    \draw[arrow] (export.east) -- (backward.west);
    \draw[arrow] (backward.east) -- (check.west);
    \draw[arrow] (check.east) -- (opt.west);
    
    % Draw arrow from Optimizer to Stream, and create a branch point
    \draw[arrow] (opt.east) -- (stream.west);
    
    % Top inputs downwards
    \draw[arrow] (loss.south) -- (backward.north);
    \draw[arrow] (opt_cfg.south) -- (opt.north);
    \draw[arrow] (accel_cfg.south) -- (stream.north);

    % Branching to Training ONNX
    % Draws a line from the calculated branch point straight down to the node
    \draw[arrow] (branch_pt) -- (train.north);
    % To make the T-junction look solid, draw a small line segment
    \filldraw[black!70] (branch_pt) circle (1.5pt);

    % Right Outputs
    \draw[arrow] (stream.east) -- (latency.west);
    \draw[arrow] (stream.east) -- (schedule.west);

\end{tikzpicture}
\caption{Overview of the MONET workflow for DNN training, enabling scheduling and latency/energy efficiency analysis on heterogeneous dataflow accelerators (HDAs). Green blocks correspond to PyTorch, blue to ONNX, and yellow to Stream. %\as{Text is too small.}
}
\label{fig:framework}
\end{figure*}
\begin{figure}[t!]
    \centering
    \includegraphics[width=\linewidth]{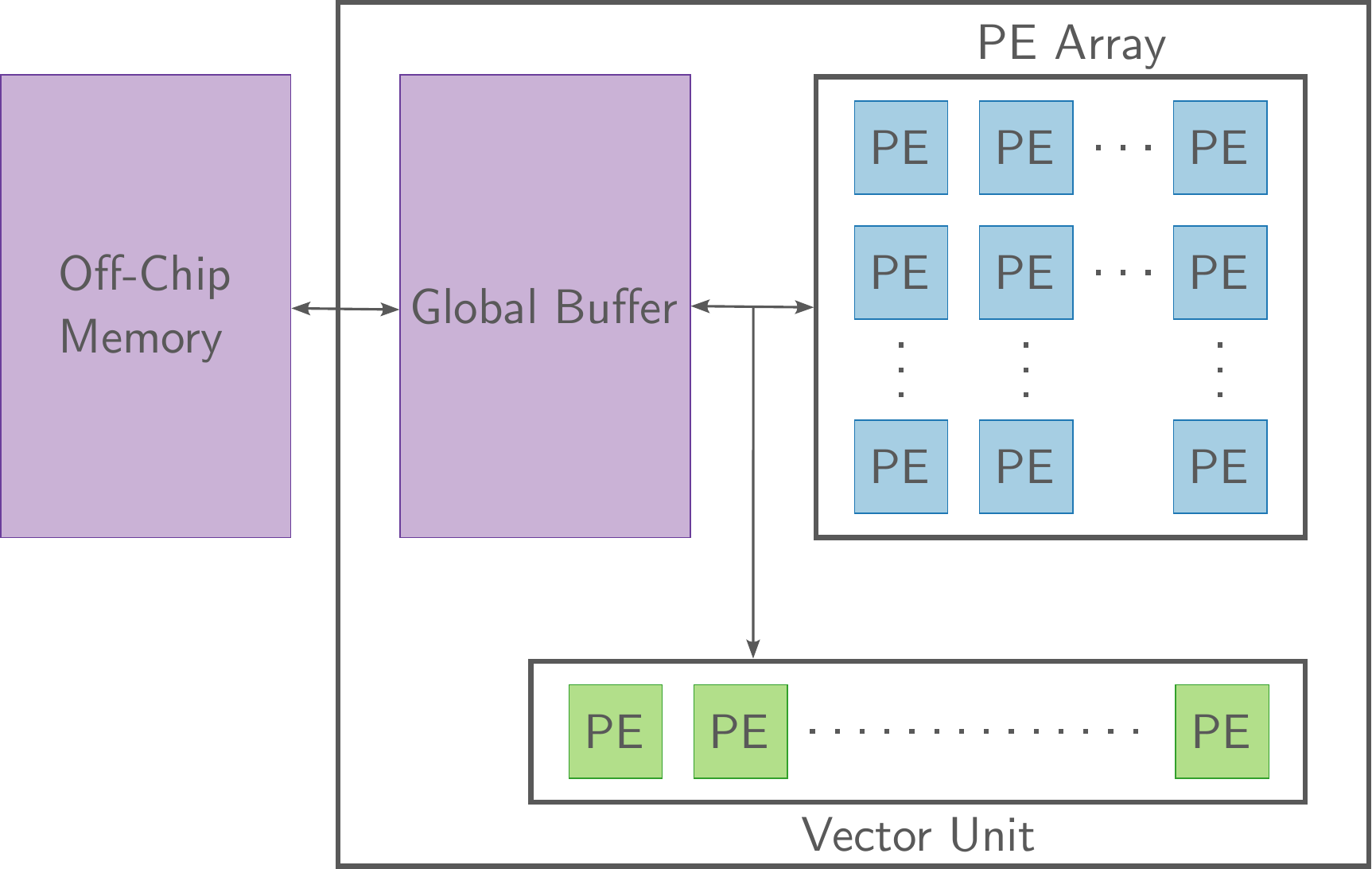}
    \caption{Stream HDA model of the FuseMax Accelerator~\cite{fusemax2024}. % \as{Text is too small.}
    }
    \label{fig:fusemax}
\end{figure}

These parallelism techniques are rarely used in isolation. Instead, they are jointly optimized to balance computational load, memory constraints, and communication costs. Achieving optimal performance on modern accelerators thus requires careful co-design of algorithms, scheduling policies, and hardware-aware optimizations.

\subsubsection{Layer-Fused Scheduling}
Layer-fused scheduling is a set of techniques that leverages the cross-layer opportunities inherent in deep neural networks to enhance performance, reduce latency, and lower energy and memory consumption. This approach operates by computing only partial feature maps of a layer and immediately utilizing these partial results for subsequent layers. 
DNNFusion~\cite{dnnfusion2021} classifies deep learning operators into mapping types (One-to-One, One-to-Many, Many-to-Many, Reorganize, Shuffle) based on the relationship between input and output elements. Fusion opportunities are analyzed using a compatibility framework that categorizes operator pairs as profitable, unprofitable, or requiring profiling based on their mapping types. The operators are then merged until no more fusion is profitable.

A notable application of layer-fused scheduling is Flash Attention~\cite{dao2022flashattention, dao2023flashattention2}, which optimizes the Softmax operation in Transformer architectures through layer fusion. In training scenarios, Flash Attention achieves up to 3.5x speedup compared to standard PyTorch Attention, demonstrating its significance for both inference and training workloads. This method not only improves computational efficiency but also reduces memory overhead, underscoring the broader impact of layer-fused scheduling in deep learning optimization.
\subsection{Modeling Frameworks}
As previously described, finding the correct hardware design for a workload and its mapping is a complex task. Most modeling frameworks used in neural network accelerator design space exploration focus on the inference part, which already represents a complex challenge. We are seeking a framework that allows for the extension of operators and mechanisms to support training workloads. Its large design space is key to discovering the best solutions.

Timeloop~\cite{timeloop} and Accelergy~\cite{accelergy} are two complementary frameworks that can be combined to systematically explore the design space of DNN accelerators. Timeloop focuses on architectural modeling and mapping: it constructs the mapping space  for a given hardware architecture by describing compute units, memory hierarchies, and interconnects, together with constraints such as dataflow and bandwidth. Its mapper then searches this space to identify efficient mappings using analytical performance models.
Accelergy provides a hierarchical and technology-aware methodology for estimating the energy consumption of hardware components. When integrated, Timeloop leverages Accelergy’s energy models to jointly evaluate performance and energy across diverse architectural configurations and mapping strategies, enabling comprehensive design space exploration.

Frameworks like ZigZag~\cite{mei2021ZigZag} expand the design space by enabling uneven operand mapping across different memory levels. However, most of these frameworks do not take into account more complex scheduling such as layer fusion and do not model the cost of data movement between operators. As previously detailed, taking into account these possibilities can be important for modeling and optimizing DNNs training workload, as the scale of computations and data movement can be a lot larger than inference workloads.

In contrast, Stream~\cite{symons2024stream} introduces a comprehensive design space exploration framework tailored for layer-fused DNNs on HDA architectures. Stream models the accelerator as a heterogeneous graph of cores, each with distinct dataflows and communication links, enabling a broad exploration of hardware configurations. By analyzing data dependencies across multiple layers, Stream determines the optimal partitioning of a DNN across the cores using a constraint optimization solver, ensuring efficient workload distribution across cores.

Stream adopts ONNX as its input format, providing native support for a wide range of DNN architectures and simplifying the implementation of custom operators. The framework parses the ONNX graph to extract operator types, specifications, and loop dimensions. This information is coupled with a mapping configuration that defines intra and inter-core partitioning, granting the user explicit control over spatial and temporal parallelism. Furthermore, Stream’s accuracy has been experimentally validated across multiple hardware platforms, ensuring its reliability in modeling real-world system performance.

In addition to academic frameworks, NVArchSim~\cite{nvarchsim} is an industrial-grade, system-level GPU simulator developed at NVIDIA. NVArchSim adopts a hybrid trace- and execution-driven methodology and prioritizes high simulation speed through carefully selected microarchitectural fidelity. By employing loosely cycle-accurate models and component-level abstraction, NVArchSim enables the simulation of large-scale HPC and machine learning workloads, including multi-GPU systems, with turnaround times suitable for architectural exploration. While NVArchSim supports both inference and training workloads and captures detailed GPU memory and interconnect behavior, it primarily targets homogeneous GPU and multi-GPU architectures rather than heterogeneous accelerator systems.

Table~\ref{table:comparisonFrameworks} summarizes and compares the main characteristics of existing DNN accelerator modeling frameworks in terms of training support, scheduling granularity, and target architectures. While prior approaches largely focus on operator-level inference modeling, none simultaneously address fine-grained layer fusion and full training workloads on heterogeneous DNN accelerators.

\begin{table}[b]
\centering
\caption{Comparison of DNN Accelerator Design Frameworks}
\begin{tabular}{|c|c|c|c|} 
\hline
\textbf{Framework}& \textbf{Training}& \textbf{Granularity} & \textbf{Target}\\ 
\hline
Timeloop~\cite{timeloop} & \multirow{2}{*}{No} & \multirow{2}{*}{Operator level} & \multirow{2}{*}{DA}\\ 
+ Accelergy~\cite{accelergy} & & &  \\
\hline
ZigZag~\cite{mei2021ZigZag}& No& Operator level &DA\\ 
\hline
\multirow{2}{*}{Dace-AD~\cite{boudaoud2025dace}} & Forward & \multirow{2}{*}{Operator level}& \multirow{2}{*}{CPU, GPU}\\
 & + Backward &  & \\
\hline
\multirow{2}{*}{Stream~\cite{symons2024stream}} & \multirow{2}{*}{No} & Fine-Grained & \multirow{2}{*}{HDA}\\
 & &  Layer Fusion & \\
\hline
\multirow{2}{*}{NVArchSim~\cite{nvarchsim}} & \multirow{2}{*}{Yes} & Warp instruction  & \multirow{2}{*}{GPU, multi-GPU} \\
 &  & level &  \\
\hline
\multirow{2}{*}{MONET(Ours)} & \multirow{2}{*}{Yes} & Fine-Grained & \multirow{2}{*}{HDA}\\
 & &  Layer Fusion & \\
\hline
\end{tabular}
\label{table:comparisonFrameworks}
\end{table}

\section{Proposed Framework}\label{lb:framework}
We extend Stream~\cite{symons2024stream} beyond its original inference-oriented design to enable accurate modeling of end-to-end neural network training on heterogeneous accelerators. Since Stream natively operates on ONNX as its primary neural network input format, our approach introduces a set of ONNX transformation passes that generate a complete training graph—including forward pass, backward pass, and optimizer updates—fully compatible with Stream. Importantly, these passes are modular and reusable, and can therefore be integrated into other ONNX-based workflows beyond Stream.

Supporting training requires both workflow-level extensions and structural enhancements to Stream’s operator representation. Training workloads introduce operators and dataflows absent from inference. To address this, we expand Stream’s operator library to support training-critical primitives such as ConvTranspose, and we introduce explicit modeling of gradient-specific data transformations, including tensor transpositions, reshaping patterns, accumulation buffers, and reduction operations arising during backpropagation. These additions allow Stream to faithfully capture the computational and memory behavior unique to training.

The workflow (Figure~\ref{fig:framework}) begins by exporting a PyTorch model to ONNX. Using ONNX Runtime Training, we generate a graph containing both forward and backward computational subgraphs. However, several ONNX gradient operators (\emph{e.g.}, ConvGrad, SoftmaxGrad) encapsulate multi-output composite computations that are incompatible with Stream’s original abstraction. We therefore introduce dedicated ONNX passes that decompose these operators into fine-grained primitives corresponding to individual gradient components (\emph{e.g.}, input, weight, and bias gradients). This decomposition enables precise scheduling, mapping, and fusion analysis within Stream.

We further integrate optimizer steps, such as SGD and ADAM, directly into the ONNX graph representation. As a result, the transformed graph captures the complete training iteration, allowing Stream to estimate latency and energy consumption while generating execution schedules for a given hardware, mapping, and fusion configuration.

To enable principled exploration of memory–computation trade-offs, we implement activation checkpointing as an additional ONNX transformation pass. Selected activations are replaced by recomputation subgraphs containing only the minimal operators and intermediate tensors required for regeneration, enabling systematic evaluation of checkpoint placement strategies.

Overall, the proposed framework preserves modularity at every stage: model specification remains in PyTorch; reusable ONNX passes generate a Stream-compatible full training graph; and hardware, mapping, and fusion parameters are provided independently for accelerator-specific performance and energy evaluation.

\section{Training Design Space Exploration}\label{lb:SearchSpace}

A large complexity characterizes the design space for training neural networks, as it encompasses multiple interdependent search spaces: the hardware architecture space, the mapping space, the layer-fusion space, and the training implementation space. Each of these dimensions introduces unique challenges, and their interactions further amplify the complexity of optimizing training workflows.

To illustrate this complexity, we analyze two representative models: ResNet-18 for image classification and GPT-2 for natural language processing (NLP). For each model, we explore variations in hardware configurations. Using our extended Stream framework, we evaluate key performance metrics such as latency and energy consumption for a single training iteration. We also generate the optimal schedule of this workload that would be implemented on the hardware. 

\subsection{Image Classification}\label{pag:resnet18}
For computer vision applications, ResNet is a well-established architecture commonly employed as a backbone for tasks such as image classification and semantic segmentation. Comparable convolutional neural network (CNN) architectures have also demonstrated efficacy for edge training in MCUNetv3 \cite{lin2022device}. 
In this part, we focus on a ResNet18 model deployed on an Edge TPU platform, as depicted in Figure~\ref{fig:edgetpuarchitecture}. We select a base ResNet18 and we model its latency and energy cost on a standard CIFAR-10 image size that is (3, 32, 32).

% \paragraph{Hardware Exploration}
\begin{figure*}[t]
\begin{center}
\def\colormap{viridis}
\def\colorData{localCompln}
\def\xName{$U \cdot L\cdot n_{PEs} \text{(log)}$}
\def\colorName{$U \cdot L$}
\def\legend{Total Computational Capacity}
\begin{tikzpicture}
    % Define the width and height for each axis
    \pgfplotsset{
        width=.48\linewidth,
        height=0.225\linewidth, % Adjusted to fit four axes vertically
    }

    % First row: Energy axes
    % Left: Energy vs. totalCompln
    \begin{scope}[xshift=-.5\linewidth, yshift=0.25\linewidth] % Adjusted yshift
        \begin{axis}[
            legend cell align={left},
            legend style={
                fill opacity=0.8,
                draw=black,
                draw opacity=1,
                text opacity=1,
                at={(0.01,0.98)},
                anchor=north west,
                draw=lightgray204
            },
            title=\it{Inference Energy and Latency},
            tick align=outside,
            tick pos=left,
            x grid style={darkgray176},
            xtick= {1.20411998265592, 1.901604224927816, 2.599088467199712, 3.296572709471608, 3.994056951743504, 4.6915411940154},
            xticklabels={16, 80, 397, 1980, 9864, 49152},
            % xticklabel style={rotate=40, anchor=north east}, % Rotate xticklabels
            scaled y ticks=base 10:-9,
            ytick scale label code/.code={},
            ylabel={Energy ($\times 10^{9}$ pJ)},
            xmajorgrids,
            xtick style={color=black},
            y grid style={darkgray176},
            xlabel=\xName,
            grid=major,
            grid style={dashed, gray!30},
            ytick style={color=black},
            ylabel near ticks,
            colormap/\colormap,
            ytick={ 9.38, 9.45, 9.53, 9.61, 9.69 },
            yticklabels={ 2, 3, 3, 4, 5 },
            point meta=explicit,
        ]
            \addplot [thick, only marks, mark=*, mark size=1.5, scatter, mark options={solid}]
                table [y=f_log_energy, x=totalCompln, meta=\colorData, col sep=comma] {figures/exploreHardware/result.csv};
            % \addplot [thick, only marks, mark=*, mark size=1.5, color=red, mark options={solid}]
            %     table [y=f_energy, x=totalCompln, col sep=comma] {figures/exploreHardware/baseline.csv};
        \end{axis}
    \end{scope}
\draw[thick, gray, dashed] (-0.075\linewidth,0.0\linewidth) -- (-0.075\linewidth,0.4\linewidth);
    % Right: Energy vs. totalCompln (second plot)
    \begin{scope}[xshift=0\linewidth, yshift=0.25\linewidth] % Adjusted yshift
        \begin{axis}[
            legend cell align={left},
            legend style={
                fill opacity=0.8,
                draw=black,
                draw opacity=1,
                text opacity=1,
                at={(0.01,0.98)},
                anchor=north west,
                draw=lightgray204
            },
            title=\it{Training Energy and Latency},
            tick align=outside,
            tick pos=left,
            x grid style={darkgray176},
            xmajorgrids,
            xtick style={color=black},
            y grid style={darkgray176},
            xlabel=\xName,
            xtick= {1.20411998265592, 1.901604224927816, 2.599088467199712, 3.296572709471608, 3.994056951743504, 4.6915411940154},
            xticklabels={16, 80, 397, 1980, 9864, 49152},
            grid=major,
            grid style={dashed, gray!30},
            ytick style={color=black},
            ylabel near ticks,
            scaled y ticks=base 10:-9,
            ytick scale label code/.code={},
            ytick={ 10.42, 10.55, 10.67, 10.79, 10.91 },
            yticklabels={ 26, 35, 47, 62, 82 },
            ylabel={Energy ($\times 10^{9}$ pJ)},
            colormap/\colormap,
            colorbar,
            colorbar style={
                title=\colorName, 
                % ytick={1.20411998265592, 1.50514997831991, 1.80617997398389, 2.10720996964787, 2.40823996531185, 2.70926996097583, 3.01029995663981},
                % yticklabels={16, 32, 64, 128, 256, 512, 1024},},
                ytick={1.20411998265592, 1.80617997398389, 2.40823996531185, 3.01029995663981},
                yticklabels={16, 64, 256, 1024},},
            point meta=explicit,
        ]
            \addplot [thick, only marks, mark=*, mark size=1.5, scatter, mark options={solid}]
                table [y=fb_log_energy, x=totalCompln, meta=\colorData, col sep=comma] {figures/exploreHardware/result.csv};
            % \addplot [thick, only marks, mark=*, mark size=1.5, color=red, mark options={solid}]
            %     table [y=fb_energy, x=totalCompln, col sep=comma] {figures/exploreHardware/baseline.csv};
        \end{axis}
    \end{scope}

    % Second row: Latency axes
    % Left: Latency vs. totalCompln
    \begin{scope}[xshift=-.5\linewidth, yshift=0.025\linewidth] % Reduced yshift
        \begin{axis}[
            legend cell align={left},
            legend style={
                fill opacity=0.8,
                draw=black,
                draw opacity=1,
                text opacity=1,
                at={(0.01,0.98)},
                anchor=north west,
                draw=lightgray204
            },
            tick align=outside,
            tick pos=left,
            x grid style={darkgray176},
            % scaled y ticks=base 10:-8,
            % ytick scale label code/.code={},
            ylabel={Latency ($\times 10^{8}$ Cycles)},
            xmajorgrids,
            xtick style={color=black},
            y grid style={darkgray176},
            xlabel=\xName,
            xtick= {1.20411998265592, 1.901604224927816, 2.599088467199712, 3.296572709471608, 3.994056951743504, 4.6915411940154},
            xticklabels={16, 80, 397, 1980, 9864, 49152},
            grid=major,
            grid style={dashed, gray!30},
            ytick style={color=black},
            ytick={ 6.62, 7.01, 7.39, 7.77, 8.15 },
            yticklabels={ 0.4, 1, 2, 6, 14 },
            ylabel near ticks,
            colormap/\colormap,
            point meta=explicit,
        ]
            \addplot [thick, only marks, mark=*, mark size=1.5, scatter, mark options={solid}]
                table [y=f_log_latency, x=totalCompln, meta=\colorData, col sep=comma] {figures/exploreHardware/result.csv};
            % \addplot [thick, only marks, mark=*, mark size=1.5, color=red, mark options={solid}]
            %     table [y=f_latency, x=totalCompln, col sep=comma] {figures/exploreHardware/baseline.csv};
        \end{axis}
    \end{scope}

    % Right: Latency vs. totalCompln (second plot)
    \begin{scope}[xshift=0.0\linewidth, yshift=0.025\linewidth] % Reduced yshift
        \begin{axis}[
            legend cell align={left},
            legend style={
                fill opacity=0.8,
                draw=black,
                draw opacity=1,
                text opacity=1,
                at={(0.01,0.98)},
                anchor=north west,
                draw=lightgray204
            },
            tick align=outside,
            tick pos=left,
            x grid style={darkgray176},
            ylabel={Latency (Cycles)},
            % scaled y ticks=base 10:-8,
            % ytick scale label code/.code={},
            ylabel={Latency ($\times 10^{8}$ Cycles)},
            xmajorgrids,
            xtick style={color=black},
            y grid style={darkgray176},
            xlabel=\xName,
            xtick= {1.20411998265592, 1.901604224927816, 2.599088467199712, 3.296572709471608, 3.994056951743504, 4.6915411940154},
            xticklabels={16, 80, 397, 1980, 9864, 49152},
            grid=major,
            grid style={dashed, gray!30},
            ytick style={color=black},
            ytick={ 7.40, 7.71, 8.02, 8.33, 8.64 },
            yticklabels={ 3, 5, 11, 21, 43 },
            ylabel near ticks,
            colormap/\colormap,
            colorbar,
            colorbar style={
                title=\colorName, 
                % ytick={1.20411998265592, 1.50514997831991, 1.80617997398389, 2.10720996964787, 2.40823996531185, 2.70926996097583, 3.01029995663981},
                % yticklabels={16, 32, 64, 128, 256, 512, 1024},},
                ytick={1.20411998265592, 1.80617997398389, 2.40823996531185, 3.01029995663981},
                yticklabels={16, 64, 256, 1024},},
            point meta=explicit,
        ]
            \addplot [thick, only marks, mark=*, mark size=1.5, scatter, mark options={solid}]
                table [y=fb_log_latency, x=totalCompln, meta=\colorData, col sep=comma] {figures/exploreHardware/result.csv};
            % \addplot [thick, only marks, mark=*, mark size=1.5, color=red, mark options={solid}]
            %     table [y=fb_latency, x=totalCompln, col sep=comma] {figures/exploreHardware/baseline.csv};
        \end{axis}
    \end{scope}
\end{tikzpicture}
\end{center}
\caption{Hardware exploration of a ResNet-18 on Edge TPU hardware for inference and training. Top row: Energy vs Accelerator Compute Ressources. Bottom row: Latency vs Accelerator Compute Ressources. Each point corresponds to a different accelerator configuration and is color-coded by its per PE Compute Resource ($U \cdot L$).}

\label{fig:resnetHWexplo}
\end{figure*}
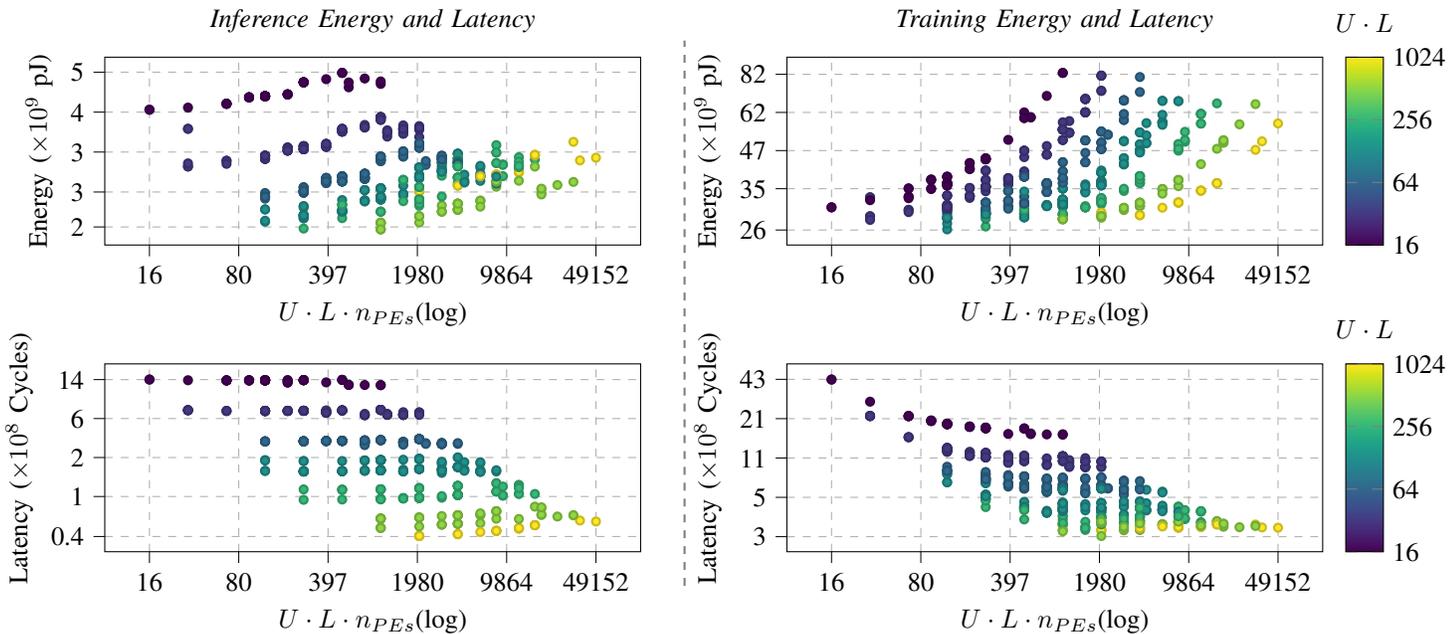

% fb_log_latency
% ytick={ 7.40, 7.71, 8.02, 8.33, 8.64 },
% yticklabels={ 3, 5, 11, 21, 43 },
% f_log_latency [6.62417294 7.00578439 7.38739584 7.76900728 8.15061873] [4.20894200e+06 1.01340814e+07 2.44003376e+07 5.87499204e+07
%  1.41455139e+08]
% ytick={ 6.62, 7.01, 7.39, 7.77, 8.15 },
% yticklabels={ 0, 1, 2, 6, 14 },
% fb_log_energy [10.42205303 10.54518458 10.66831612 10.79144766 10.91457921] [33592.329586   37994.03970748 42972.51994976 48603.34634195
%  54971.99788138]
% ytick={ 10.42, 10.55, 10.67, 10.79, 10.91 },
% yticklabels={ 26, 35, 47, 62, 82 },
% f_log_energy [9.3751731  9.45353833 9.53190355 9.61026878 9.688634  ] [11791.95859401 12753.21056481 13792.82147353 14917.17895144
%  16133.19133408]
% ytick={ 9.38, 9.45, 9.53, 9.61, 9.69 },
% yticklabels={ 2, 3, 3, 4, 5 },
We systematically explore the hardware design space of an Edge TPU heterogeneous dataflow accelerator (HDA) model, based on the configuration defined in~\cite{Zhou2022CoDesignTPU}. The baseline configuration features a $4 \times 4$ array of processing elements (PEs). Each PE is designed as a weight-stationary dataflow core, which is highly effective for accelerating convolutional neural network (CNN) workloads. The PE architecture includes 2~MBs of memory and 4 compute lanes, with each lane comprising a 32~KBs register file and 64 4-way single instruction, multiple data (SIMD) units. We explore variation of this configuration, as detailed in Table~\ref{tab:tpusearchspace}, to discover the properties of this search space and the differences between inference and training. A manually designed layer-fusion configuration for the inference is used in both inference and training to show the different trade-offs reached in this kind of configuration. 
In addition, we leverage both temporal and spatial parallelism to better exploit the architectural heterogeneity between SIMD cores and weight-stationary PEs. Specifically, pipeline parallelism is used to map layers to the most suitable compute units according to their computational characteristics, while tensor parallelism distributes convolutional output channels across multiple weight-stationary PEs to increase concurrency and improve resource utilization.
Figure~\ref{fig:resnetHWexplo} presents the outcomes of this exploration, illustrating latency and energy for both inference and training workloads in comparison with the total compute resources of each accelerator, $U \cdot L \cdot n_{PEs}$, where $U \cdot L$ represents the compute resource of each PE.

\begin{table}[b!]
    \centering
    \caption{Edge TPU search space, bold denotes the baseline configuration, $n_{PEs}=x_{PEs} \cdot y_{PEs}$}
    \begin{tabular}{c c}
         Parameter&  Search space\\ \hline
         xPEs & 1, 2, \textbf{4}, 6, 8 \\ \hline
         yPEs & 1, 2, \textbf{4}, 6, 8 \\ \hline
         SIMD units \textbf{U} & 16, 32, \textbf{64}, 128\\ \hline
         Compute Lanes \textbf{L}& 1, 2, \textbf{4}, 8\\ \hline
         Local Memory (MBs) & 0.5, 1, \textbf{2}, 3, 4\\ \hline
         Register File (KBs) & 8, 16, 32, \textbf{64}, 128\\ \hline
    \end{tabular}
    % \vspace{0.2cm} % Adjust the value (\emph{e.g.}, 0.5cm) to control the space    
    \label{tab:tpusearchspace}
\end{table}

As shown in Figure~\ref{fig:main}, the distributions of energy and latency for inference and training workloads exhibit significant structural differences. These discrepancies highlight that the hardware characteristics that favor one operational mode do not systematically benefit the other, which is a key consideration when targeting unified accelerators.

To further analyze these trade-offs, Figure~\ref{fig:resnetHWexplo} represents energy and latency as functions of the total compute resource while accounting for the compute capacity per processing element (PE). This view allows us to evaluate the impact of PE granularity under fixed resource constraints. For training latency, larger PEs do not appear on the Pareto front, indicating that increasing PE size does not translate into optimal latency-performance trade-offs. In contrast, for inference latency, larger PEs almost always reduce execution time and are strongly associated with low-latency Pareto-optimal configurations.

The energy perspective reveals a different trend. During training, larger PEs are more efficient when considering very large total compute budgets, but they are not the best designs in terms of absolute energy optimality. For inference, larger PEs are not energy-efficient and never lie on the energy Pareto front. This asymmetry between latency and energy behavior reinforces that PE scaling has workload-dependent implications.

Overall, the distribution of results differs markedly between training and inference, confirming that an architecture optimized for one objective or workload does not directly generalize to the other.

\subsection{Natural Language Processing}
At the opposite end of the computational spectrum, Large Language Models (LLMs) such as GPT-2 represent the high-performance workloads typical of cloud-scale accelerators. The workload consists of a standard Transformer architecture configured with a fixed sequence length and a causal attention mask. For the hardware architecture, we use FuseMax~\cite{fusemax2024} as a recent architecture that can easily be modeled in Stream~\cite{zigzagllm2024}. A possible HDA representation is shown in Figure~\ref{fig:fusemax}. This architecture is composed of a large array of Multiply-Accumulate (MAC) elements designed as an output-stationary dataflow accelerator and a large vector array. Both arrays' memories are linked together and they can access a large on-chip buffer that can interact with an off-chip memory. 
As the architecture consists of two distinct cores, an output-stationary dataflow core and a SIMD core, we primarily employ pipeline parallelism to enable efficient operand transfer and overlap computation between them.
We evaluate the latency and memory for this architecture with the variations presented in Table~\ref{tab:fusemaxsearchspace}. 

\begin{table}[b!]
    \centering
    \caption{FuseMax Search Space}
    \begin{tabular}{c c}
         Parameter&  Search space\\ \hline
         xPEs & 64, 128, 256, 512 \\ \hline
         yPEs & 64, 128, 256, 512 \\ \hline
         Vector PEs & 32, 64, 128, 256\\ \hline
         % Buffer Bandwidth & 2048, 4096, 8192, 16384\\ \hline
         Buffer Bandwidth & 8192, 16384\\ \hline
         Buffer Size (MBs) & 4, 8, 16, 32 \\ \hline
         Off-Chip Bandwidth & 512, 1024, 2048, 4096, 8192\\ \hline
    \end{tabular}
    \vspace{0.2cm}
    
    \label{tab:fusemaxsearchspace}
\end{table}

\begin{figure}[t!]
\begin{center}
\def\colormap{viridis}
\def\colorData{buffer_bandwidth_bpc}

\begin{tikzpicture}
\begin{groupplot}[
    group style={group size=1 by 2, vertical sep=1.5cm},
    width=0.8\linewidth,
    height=0.5\linewidth,
    tick align=outside,
    tick pos=left,
    x grid style={darkgray176},
    y grid style={darkgray176},
    xmajorgrids,
    grid=major,
    grid style={dashed, gray!30},
    xtick style={color=black},
    scaled x ticks=base 10:-9,
    xtick scale label code/.code={},
    xlabel={Energy ($\times 10^{9}$ pJ)},
    ytick style={color=black},
    ylabel near ticks,
    scaled y ticks=base 10:-7,
    ytick scale label code/.code={},
    ylabel={Latency ($\times 10^{7}$ Cycles)},
    colormap/\colormap,
    point meta=explicit,
    legend cell align={left},
    legend style={
      fill opacity=0.8,
      draw=black,
      draw opacity=1,
      text opacity=1,
      at={(0.01,0.98)},
      anchor=north west,
      draw=lightgray204
    },
]

% --- Top: Inference (NO colorbar) ---
\nextgroupplot[
    title=\it{Inference},
    colorbar,
    xlabel={},
    % xmin=2e9, xmax=4e9,
    % ymin=0, ymax=0.8e8,
    colorbar style={
        ytick style={draw=none},
        title=\shortstack{Buffer\\Bandwidth},
        title style={align=center},
        height=0.7\linewidth + 1cm,
        scaled y ticks=false, % <- do not apply scaled y ticks to colorbar
        ytick scale label code/.code={} % <- no "×10^k" scale label on the colorbar
    },
]

\addplot [thick, only marks, mark=*, mark size=1.2, scatter]
table [y=f_latency, x=f_energy, meta=\colorData,col sep=comma] {figures/SA_GPT/resultSA.csv};

% --- Bottom: Training (the ONLY one with colorbar) ---
\nextgroupplot[
    title=\it{Training},
]
\addplot [thick, only marks, mark=*, mark size=1.2, scatter]
table [y=fb_latency, x=fb_energy, meta=\colorData,col sep=comma] {figures/SA_GPT/resultSA.csv};

\end{groupplot}
\end{tikzpicture}
\end{center}
\caption{Energy–latency trade-offs of a small GPT-2 model on the FuseMax accelerator. Results are shown for inference (left) and training (right). Each point represents a distinct hardware configuration, with energy consumption (pJ) plotted against execution latency (cycles) and color-coded by buffer bandwidth (in Bytes per cycle), highlighting the impact of memory bandwidth on performance and energy efficiency.}
\label{fig:GPTHWExplo}
\end{figure}
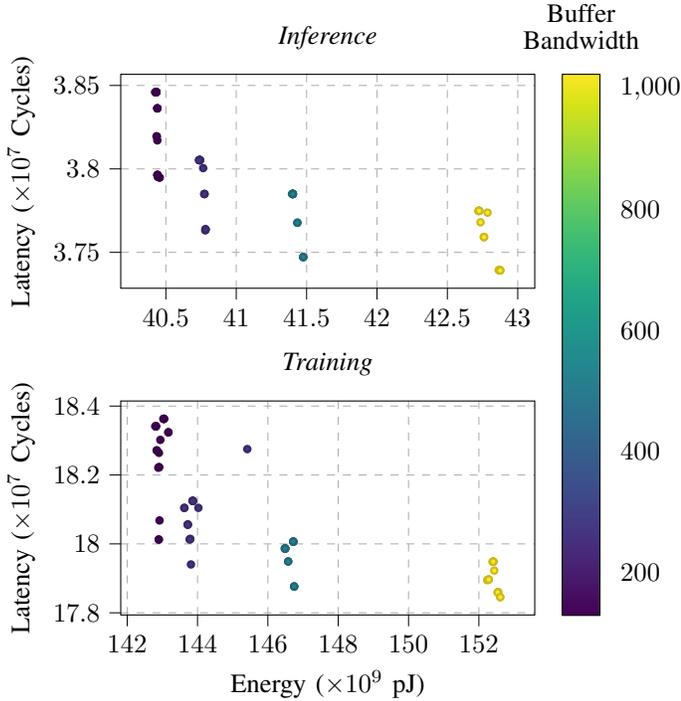

Compared to the previous scenario, both the hardware and neural network architectures have a more regular structure. The FuseMax accelerator features a more streamlined architecture relative to the Edge TPU. Similarly,  the small GPT2 model is significantly more homogeneous than the ResNet-18. It consists of a repetition of identical Attention and Feed-Forward blocks. The structural homogeneity is reflected in Figure~\ref{fig:fusemax}, where the performance distributions appear more concentrated and less sensitive to architectural variations.

While the exploration presented in Section~\ref{lb:SearchSpace} illustrates the complexity of the training design space, it also points to a practical challenge: exhaustive exploration alone does not efficiently identify optimal configurations. The interaction between hardware architecture, mapping strategies, layer-fusion choices, and training-specific mechanisms creates a large combinatorial space in which manual tuning becomes difficult.
To effectively navigate this complexity, targeted optimization strategies are required. In the next section, we therefore introduce methods that systematically reduce the search space while preserving high-quality solutions. Specifically, we focus on two optimization problems that we found to be particularly important for training-aware design: fused-layer scheduling and activation checkpointing.

\section{Optimizations}
In this part, we propose methods to efficiently explore the fused-layer search space as well as the activation checkpointing search space. 
\subsection{Layer Fusion Strategy}\label{pag:lfs}

Determining the optimal partitioning of a workload graph into fused subgraphs is a combinatorial optimization problem. While feasible for small inference graphs with around a hundred computational nodes, it becomes significantly more complex for training workloads, which typically contain several times more nodes.

Moreover, training workloads contain several properties that make layer-fusion relevant. Optimizers such as SGD or ADAM contain only element-wise operations, making them good candidates to be fused with the weight gradient computation. This fusion can reduce the memory necessary by discarding as soon as possible the saved activations.

To this end, we extend the capabilities of Stream with a new fused-layer solver based on constraint optimization.

\subsubsection{Problem Formulation}

Consider a workload graph \(\mathcal{G} = (\mathcal{V}, \mathcal{E})\) that represents a forward pass, a combined forward-backward pass, or any neural network computation. Here, \(\mathcal{V}\) denotes the set of operators (nodes), and \(\mathcal{E}\) represents the tensors (edges) exchanged between them. Given an accelerator composed of a set of cores \(\mathcal{C}\), the goal is to partition \(\mathcal{G}\) into a set of subgraphs \(\mathcal{S}\). 
For each node $i \in \mathcal{V}'$, $\mathcal{G}'=(\mathcal{V}', \mathcal{E}') \subseteq  \mathcal{G}$, and core $c \in \mathcal{C}$, we define:
\begin{itemize}
    \item $m_{i,c}$: memory required by node $i$ on core $c$;
    \item $T_i$: intra-core tiling factor of node $i$;
    \item $M_c$: available local memory on core $c$.
\end{itemize}

To determine candidate subgraphs, we perform a Breadth-First Search (BFS) from each node in $\mathcal{G}$. Since this search explores all possible BFS paths from each node, it is a combinatorial problem whose complexity grows exponentially with the number of nodes. We limit the maximum length of the BFS to make the search tractable. We apply the following constraints as a backtracking approach to reduce the complexity of the search: 
\paragraph*{Memory Constraint}
The total memory consumed by all nodes assigned to core $c$ must not exceed its available capacity:
\[
\sum_{i \in \mathcal{V}} m_{i,c} \leq M_c, \quad \forall c \in \mathcal{C}
\]

\paragraph*{Intra-Core Tiling Constraint}
For each operator in the DNN model, a subset of the loops defining its computation is designated as outer temporal loops and expressed as intra-core tilings in Stream. These tilings are used to validate operator fusion. This requirement imposes an additional constraint: within a given subgraph, all intra-core tiling factors must be mutually compatible, such that each tiling factor divides every other tiling factor in the set.

\[
\forall i, j \in \mathcal{V}, \quad T_i \mid T_j \quad \text{or} \quad T_j \mid T_i
\]

\paragraph*{Operator Type Constraint}
To limit the complexity of scheduling highly complex fused subgraphs, we impose a constraint on the number of convolutional and general matrix multiplication (GEMM) operations. Specifically, each subgraph is restricted to containing no more than three convolutional operations and no more than two GEMM operations. This restriction ensures that the analysis cost to generate a valid schedule for the fused operators in the subgraph is not too high. This is similar to the constraints applied in DNNFusion~\cite{dnnfusion2021}

Following this search, we apply the following constraint to each subgraph: 
\[
\sum_{v \in \mathcal{V}_g} o_v \leq 1
\]
where $\mathcal{V}_g$ is the set of nodes in subgraph $g$, and $o_v$ is an indicator function such that $o_v = 1$ if node $v$ has outgoing edges, and $0$ otherwise. This ensures that the resulting fused subgraph in Stream does not produce intermediate tensors required by other subgraphs, thereby avoiding off-chip tensor storage. 

Finding the optimal combination of valid subgraphs that fully cover $\mathcal{G}$ is combinatorial in nature, with complexity $\mathcal{O}(2^N)$, making exhaustive evaluation infeasible for practical networks (\emph{e.g.}, $N \approx 500$ for ResNet-18 training). We therefore employ Integer Programming (IP) with a heuristic goal to approximate the best solution. 

We define the binary decision variable:
\[
x_g \in \{0,1\}, \quad g \in \mathcal{S}
\]
where $x_g = 1$ if subgraph $g$ is selected in the final partition, and $0$ otherwise.

The optimization objective minimizes the total number of selected subgraphs in order to maximize the fusion opportunities while subject to full node coverage:
\[
\begin{aligned}
&\text{Minimize} \quad  \sum_{g \in \mathcal{S}} x_g, & \text{ subject to} \quad  \sum_{g \in \mathcal{S}: i \in g} x_g = 1, \quad \forall i \in \mathcal{V}
\end{aligned}
\]

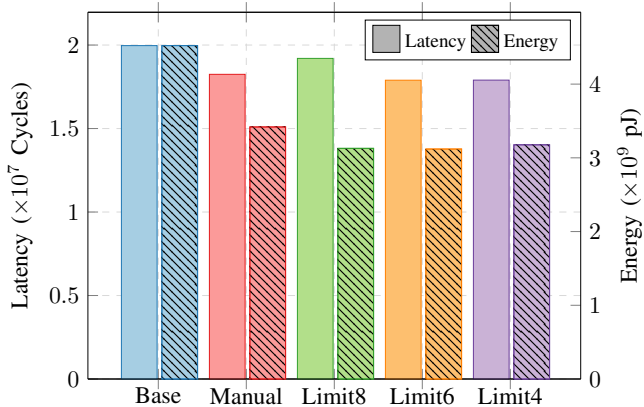
\begin{figure}[ht!]
    \centering
    \resizebox{\linewidth}{!}{% \documentclass{article}
% \usepackage{pgfplots}
% \pgfplotsset{compat=1.18}
% \usetikzlibrary{patterns} % Required for patterns
% \input{colors}
\pgfmathsetmacro{\shift}{-8pt}
% \begin{document}
\begin{tikzpicture}
\begin{axis}[
    width=\columnwidth,
    height=7cm,
    bar width=15pt,
    enlarge x limits=0.2,
    symbolic x coords={Base, Manual, Limit8, Limit6, Limit4},
    xtick={Base,Manual,Limit8, Limit6, Limit4},        % <-- IMPORTANT
    ylabel={Latency (Cycles)},
    axis y line*=left,
    ymin=0,
    scaled y ticks=base 10:-7,
    ytick scale label code/.code={},
    ylabel={Latency ($\times 10^{7}$ Cycles)},
    grid=major,
    grid style={dashed, gray!30},
]

\addplot[fill=Paired-2, draw=Paired-1, ybar, bar shift=-8pt]
    coordinates {(Base, 19966170.0)};
\addplot[fill=Paired-6, draw=Paired-5, ybar, bar shift=-8pt]
    coordinates {(Manual, 18246096.0)};
\addplot[fill=Paired-4, draw=Paired-3, ybar, bar shift=-8pt]
    coordinates {(Limit8, 19200693.0)};
\addplot[fill=Paired-8, draw=Paired-7, ybar, bar shift=-8pt]
    coordinates {(Limit6,  17894203.0)};
\addplot[fill=Paired-10, draw=Paired-9, ybar, bar shift=-8pt]
    coordinates {(Limit4, 17899628)};
% \addlegendentry{Latency}
\end{axis}

\begin{axis}[
ybar,
    width=\columnwidth,
    height=7cm,
    bar width=15pt,
    enlarge x limits=0.2,
    symbolic x coords={Base, Manual, Limit8, Limit6, Limit4},
    xtick={Base,Manual,Limit8, Limit6, Limit4},        % <-- IMPORTANT
    ylabel={Energy (pJ)},
    xticklabels = {},
    ymin=0,
    scaled y ticks=base 10:-9,
    ytick scale label code/.code={},
    ylabel={Energy ($\times 10^{9}$ pJ)},
    % ymin=-2e9, ymax=4e9,
    % ymin = 15,
    axis y line*=right,
    legend columns=2,
    legend style={
        at={(0.98,0.98)},
        anchor=north east,
        font=\footnotesize,
    },
    % legend image code/.code={\draw[fill=Set2-8, draw=Set1-10, postaction={pattern=north west lines}] (0cm,0cm) rectangle (0.4cm,0.4cm);}
]
\addlegendimage{
    legend image code/.code={
        \draw[fill=Set2-8, draw=Set1-10]
            (0cm,-0.15cm) rectangle (0.4cm,0.25cm);
    }
}
\addlegendentry{Latency}

% ---- Energy legend (patterned square)
\addlegendimage{
    legend image code/.code={
        \draw[fill=Set2-8, draw=Set1-10, postaction={pattern=north west lines}]
            (0cm,-0.15cm) rectangle (0.4cm,0.25cm);
    }
}
\addlegendentry{Energy}
\addplot[
    fill=Paired-2,
    draw=Paired-1,
    bar shift=17pt + \shift, ybar,
    postaction={pattern=north west lines}
] coordinates {(Base, 4522230546.616)};

\addplot[
    fill=Paired-6,
    draw=Paired-5,
    bar shift=17pt + \shift, ybar,
    postaction={pattern=north west lines}
] coordinates {(Manual, 3418727585.01598)};
\addplot[
    fill=Paired-4,
    draw=Paired-3,
    bar shift=17pt + \shift, ybar,
    postaction={pattern=north west lines}
] coordinates {(Limit8, 3128449313.3809853)};
\addplot[
    fill=Paired-8,
    draw=Paired-7,
    bar shift=17pt + \shift, ybar,
    postaction={pattern=north west lines}
] coordinates {(Limit6, 3119411313.380985)};
\addplot[
    fill=Paired-10,
    draw=Paired-9,
    bar shift=17pt + \shift, ybar,
    postaction={pattern=north west lines}
] coordinates {(Limit4, 3176560031.7559857)};
\end{axis}
\end{tikzpicture}
% \end{document}
}
    \caption{Latency and energy for ResNet18 inference on an Edge TPU using different layer-fusion strategies. Base is the layer-by-layer approach, Manual is a manually designed fusion, and Limit4 to 8 are our algorithm with varying subgraph limits. Our approach reduces both latency and energy compared to manual fusion, enabling faster workflow iteration.}
    \label{fig:layerfusion}
\end{figure}

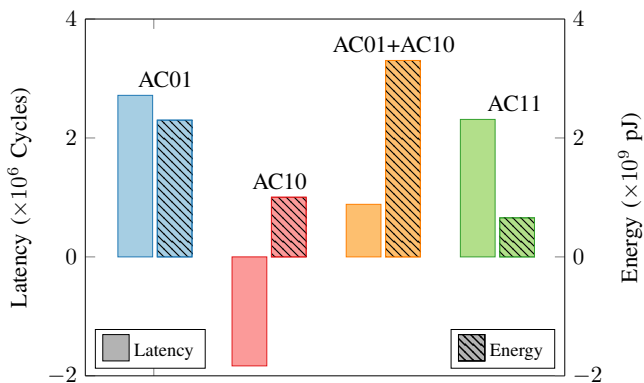
\begin{figure}[hb!]
\centering
\resizebox{\linewidth}{!}{% \documentclass{article}
% \usepackage{pgfplots}
% \pgfplotsset{compat=1.18}
% \usetikzlibrary{patterns} % Required for patterns
% \input{colors}
\pgfmathsetmacro{\shift}{-8pt}
% \begin{document}
\begin{tikzpicture}
\begin{axis}[
    width=\columnwidth,
    height=7cm,
    bar width=15pt,
    enlarge x limits=0.2,
    symbolic x coords={AC01, AC10, AC2, AC11},
    xtick=data,
    xticklabels={},
    ylabel={Latency (Cycles)},
    ymin=-2e6, ymax=40e5,
    axis y line*=left,
    scaled y ticks=base 10:-6,
    ytick scale label code/.code={},
    ylabel={Latency ($\times 10^{6}$ Cycles)},
    legend style={at={(0.02,0.02)}, anchor=south west, font=\footnotesize},
    legend image code/.code={\draw[fill=Set2-8, draw=Set1-10] (0cm,-0.15cm) rectangle (0.4cm,0.25cm);}
]

\addplot[fill=Paired-2, draw=Paired-1, ybar, bar shift=-8pt] coordinates {(AC01, 2716606)};
\node[above, xshift=5pt] at (axis cs:AC01, 2716606) {AC01};
\addplot[fill=Paired-6, draw=Paired-5, ybar, bar shift=-8pt] coordinates {(AC10, -1832621.66666663)};
\addplot[fill=Paired-8, draw=Paired-7, ybar, bar shift=-8pt] coordinates {(AC2, 883985.33333334)};
\addplot[fill=Paired-4, draw=Paired-3, ybar, bar shift=-8pt] coordinates {(AC11, 2311836.33)};
\node[above, xshift=8pt] at (axis cs:AC11,2311836.33) {AC11};
\addlegendentry{Latency}
\end{axis}

\begin{axis}[
ybar,
    width=\columnwidth,
    height=7cm,
    bar width=15pt,
    enlarge x limits=0.2,
    symbolic x coords={AC01, AC10, AC2, AC11},
    xtick=data,
    ylabel={Energy (pJ)},
    ymin=-2e9, ymax=4e9,
    axis y line*=right,
    axis x line=none,
    scaled y ticks=base 10:-9,
    ytick scale label code/.code={},
    ylabel={Energy ($\times 10^{9}$ pJ)},
    legend style={at={(0.98,0.02)}, anchor=south east, font=\footnotesize},
    legend image code/.code={\draw[fill=Set2-8, draw=Set1-10, postaction={pattern=north west lines}] (0cm,-0.15cm) rectangle (0.4cm,0.25cm);}
]

% Stripes for Energy AC01
\addplot[
    fill=Paired-2,
    draw=Paired-1,
    bar shift=17pt + \shift, ybar,
    postaction={pattern=north west lines}
] coordinates {(AC01, 2299541203.14)};
% Manually place the node for AC01

% Stripes for Energy AC10
\addplot[
    fill=Paired-6,
    draw=Paired-5,
    bar shift=17pt + \shift, ybar,
    postaction={pattern=north west lines}
] coordinates {(AC10, 1003261192.06)};
% Manually place the node for AC10
\node[above, xshift=5pt] at (axis cs:AC10, 1003261192.06) {AC10};

% Stripes for Energy AC01+AC10
\addplot[
    fill=Paired-8,
    draw=Paired-7,
    bar shift=17pt + \shift, ybar,
    postaction={pattern=north west lines}
] coordinates {(AC2, 3302802395.12)};
% Manually place the node for AC01+AC10
\node[above, xshift=5pt] at (axis cs:AC2, 3302802395.12) {AC01+AC10};
\addlegendentry{Energy}
% Stripes for Energy AC11
\addplot[
    fill=Paired-4,
    draw=Paired-3,
    bar shift=17pt + \shift, ybar,
    postaction={pattern=north west lines}
] coordinates {(AC11, 656826923.12)};
% Manually place the node for AC11
\end{axis}
\end{tikzpicture}
% \end{document}
}
\caption{ResNet18 Performance Under Activation Checkpointing: Energy and Latency Analysis.
All values are relative to the baseline (saving all activations). Striped bars show energy consumption, while solid bars highlight latency for each strategy.}
\label{fig:resnetnonlinearity}
\end{figure}

\begin{figure*}[h!t]
\centering
\resizebox{\linewidth}{!}{% \begin{figure*}[h!]
% \begin{center}
\begin{tikzpicture}
\begin{scope}[ xshift=-.5\linewidth]
\pgfplotsset{%
    width=.5\linewidth,
    height=0.4\linewidth
}
        \begin{axis}[
        legend cell align={left},
        legend style={
          fill opacity=0.8,
          draw=black,
          draw opacity=1,
          text opacity=1,
          at={(0.01,0.98)},
          anchor=north west,
          draw=lightgray204
        },
        tick align=outside,
        tick pos=left,
        x grid style={darkgray176},
        xlabel={Saved Memory (\%)},
        xmajorgrids,
        xtick style={color=black},
        y grid style={darkgray176},
        ylabel={Energy (pJ)},
        scaled y ticks=base 10:-11,
        ytick scale label code/.code={},
        ylabel={Energy ($\times 10^{11}$ pJ)},
        grid=major,
        grid style={dashed, gray!30},
        ytick style={color=black},
        ylabel near ticks,
        ]

    \addplot [
        only marks, % Use either 'only marks' or 'scatter', not both
        mark=*,
        mark size=2,
        color=Paired-4,
        mark options={solid, fill=Paired-10, draw=Paired-9}
    ] table [x=PSavedMemory0, y=Energy0, col sep=comma] {figures/GA/output.csv};
    \addlegendentry{Gen $0$}
    \addplot [
        only marks, % Use either 'only marks' or 'scatter', not both
        mark=*,
        mark size=2,
        color=Paired-4,
        mark options={solid, fill=Paired-2, draw=Paired-1}
    ] table [x=PSavedMemory1, y=Energy1, col sep=comma] {figures/GA/output.csv};
    \addlegendentry{Gen $1$}
    \addplot [
        only marks, % Use either 'only marks' or 'scatter', not both
        mark=*,
        mark size=2,
        color=Paired-4,
        mark options={solid, fill=Paired-6, draw=Paired-5}
    ] table [x=PSavedMemory2, y=Energy2, col sep=comma] {figures/GA/output.csv};
    \addlegendentry{Gen $2$}
    \addplot [
        only marks, % Use either 'only marks' or 'scatter', not both
        mark=*,
        mark size=2,
        color=Paired-4,
        mark options={solid, fill=Paired-8, draw=Paired-7}
    ] table [x=PSavedMemory3, y=Energy3, col sep=comma] {figures/GA/output.csv};
    \addlegendentry{Gen $3$}
    \addplot [
        only marks, % Use either 'only marks' or 'scatter', not both
        mark=*,
        mark size=2.5,
        color=Paired-4,
        mark options={solid, fill=Paired-4, draw=Paired-3}
    ] table [x=BaselineMem, y=BaselineEnergy, col sep=comma] {figures/GA/output.csv};
    \addlegendentry{Baseline}
    % \addplot [
    %     only marks, % Use either 'only marks' or 'scatter', not both
    %     mark=*,
    %     mark size=2,
    %     color=Paired-4,
    %     mark options={solid, fill=Paired-4, draw=Paired-3}
    % ] table [x=PSavedMemory4, y=Energy4, col sep=comma] {figures/GA/output.csv};
    % \addlegendentry{Gen $4$}
    % \addplot [
    %     only marks, % Use either 'only marks' or 'scatter', not both
    %     mark=*,
    %     mark size=2,
    %     color=Paired-4,
    %     mark options={solid, fill=Paired-12, draw=Paired-11}
    % ] table [x=PSavedMemory5, y=Energy5, col sep=comma] {figures/GA/output.csv};
    % \addlegendentry{Gen $5$}

\end{axis}
\end{scope}
\begin{scope}
\pgfplotsset{%
    width=.5\linewidth,
    height=0.4\linewidth
}
        \begin{axis}[
        legend cell align={left},
        legend style={
          fill opacity=0.8,
          draw=black,
          draw opacity=1,
          text opacity=1,
          at={(0.01,0.98)},
          anchor=north west,
          draw=lightgray204
        },
        tick align=outside,
        tick pos=left,
        x grid style={darkgray176},
        xlabel={Saved Memory (\%)},
        xmajorgrids,
        xtick style={color=black},
        y grid style={darkgray176},
        ylabel={Latency (Cycles)},
        scaled y ticks=base 10:-8,
        ytick scale label code/.code={},
        ylabel={Latency ($\times 10^{8}$ Cycles)},
        grid=major,
        grid style={dashed, gray!30},
        ytick style={color=black},
        ylabel near ticks,
        ]

        % \addplot[fill=Paired-2, draw=Paired-1]   coordinates {(0.75, 2716607.33333334)};
        % \addplot[fill=Paired-6, draw=Paired-5]    coordinates {(1, -1832621.66666663)};
        % \addplot[fill=Paired-8, draw=Paired-7]  coordinates {(1.25, 2311836.33333334)};
        % \addplot[fill=Paired-4, draw=Paired-3] coordinates {(1.5, 883985.666666716)};

    \addplot [
        only marks, % Use either 'only marks' or 'scatter', not both
        mark=*,
        mark size=2,
        color=Paired-4,
        mark options={solid, fill=Paired-10, draw=Paired-9}
    ] table [x=PSavedMemory0, y=Latence0, col sep=comma] {figures/GA/output.csv};
    \addlegendentry{Gen $0$}
    \addplot [
        only marks, % Use either 'only marks' or 'scatter', not both
        mark=*,
        mark size=2,
        color=Paired-4,
        mark options={solid, fill=Paired-2, draw=Paired-1}
    ] table [x=PSavedMemory1, y=Latence1, col sep=comma] {figures/GA/output.csv};
    \addlegendentry{Gen $1$}
    \addplot [
        only marks, % Use either 'only marks' or 'scatter', not both
        mark=*,
        mark size=2,
        color=Paired-4,
        mark options={solid, fill=Paired-6, draw=Paired-5}
    ] table [x=PSavedMemory2, y=Latence2, col sep=comma] {figures/GA/output.csv};
    \addlegendentry{Gen $2$}
    \addplot [
        only marks, % Use either 'only marks' or 'scatter', not both
        mark=*,
        mark size=2,
        color=Paired-4,
        mark options={solid, fill=Paired-8, draw=Paired-7}
    ] table [x=PSavedMemory3, y=Latence3, col sep=comma] {figures/GA/output.csv};
    \addlegendentry{Gen $3$}
    \addplot [
        only marks, % Use either 'only marks' or 'scatter', not both
        mark=*,
        mark size=2.5,
        color=Paired-4,
        mark options={solid, fill=Paired-4, draw=Paired-3}
    ] table [x=BaselineMem, y=BaselineLatence, col sep=comma] {figures/GA/output.csv};
    \addlegendentry{Baseline}

    % 213577542.6666627
    % \addplot [
    %     only marks, % Use either 'only marks' or 'scatter', not both
    %     mark=*,
    %     mark size=2,
    %     color=Paired-4,
    %     mark options={solid, fill=Paired-4, draw=Paired-3}
    % ] table [x=PSavedMemory4, y=Latence4, col sep=comma] {figures/GA/output.csv};
    % \addlegendentry{Gen $4$}
    % \addplot [
    %     only marks, % Use either 'only marks' or 'scatter', not both
    %     mark=*,
    %     mark size=2,
    %     color=Paired-4,
    %     mark options={solid, fill=Paired-12, draw=Paired-11}
    % ] table [x=PSavedMemory5, y=Latence5, col sep=comma] {figures/GA/output.csv};
    % \addlegendentry{Gen $5$}
        % \addplot [thick,
        % mark=*,
        % mark size=2,
        % color=red,
        % mark options={solid}] table [x=fb_energy, y=fb_Latence,col sep=comma]   {figures/exploreHardware/pareto_fb.csv};
        % \addlegendentry{Pareto Front}
        \end{axis}

\end{scope}
\end{tikzpicture}
% \end{center}
% \caption{Genetic algorithm activation checkpointing.}
% \label{fig:GA_AC}
% \end{figure*}
}
\caption{Optimization of activation checkpointing for ResNet-18 training (Adam optimizer, batch size 1, 224$\times$224 inputs) using a genetic algorithm. The reported memory savings correspond to the percentage of the total activation memory that is avoided by recomputing intermediate tensors during the backward pass instead of storing them.}
\label{fig:GA_AC}
\end{figure*}
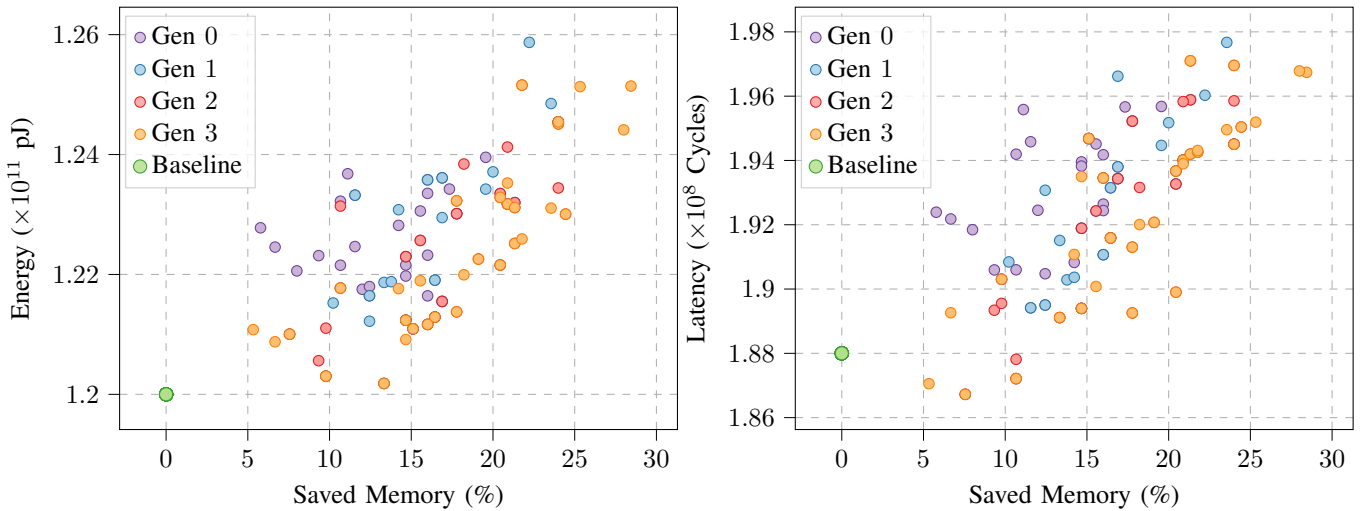

Alternative heuristics, such as minimizing inter-subgraph tensor sizes, can also be considered to further reduce memory overhead. However, they can be very expensive, especially when considering training workloads, because the complexity grows from linear to quadratic.

\subsubsection{Results}

To validate our algorithm, we evaluate its performance on ResNet-18 inference using an Edge TPU. We compare our results to both baseline layer-by-layer scheduling and the manual fusion configurations from Stream. We also investigate how the maximum subgraph length influences fusion efficiency.
As illustrated in Figure~\ref{fig:layerfusion}, layer-fusion proves to be a highly effective technique for minimizing both latency and energy consumption in DNN model inference. Our algorithm outperforms the baseline layer-by-layer approach and most of the time the manually designed fusion configuration, achieving notable reductions in both metrics.

The results reveal that the optimal performance—when operator type constraints are not enforced—is achieved with a maximum subgraph limit of 6, with a length of 4 having a similar latency. This finding underscores the efficiency of our approach, enabling users to accelerate their workflow iterations without the need to manually reconfigure layer-fusion settings for each different workload, mapping and hardware configurations.

\subsection{Activation Checkpointing}\label{pag:AC}

\label{subsec:activation_checkpointing}

As previously described, activation checkpointing expands the algorithmic search space during training by determining whether to checkpoint or recompute activations required for the backward pass. In this section, we focus on exploring the activation checkpointing search space and demonstrate the effectiveness of a genetic algorithm in efficiently navigating this space, particularly for layer-fused neural networks.

\subsubsection{Linear Model Limits}
A key challenge arises from the inadequacy of the Mixed-Integer Linear Programming (MILP) model in representing fused-layer neural networks. Specifically, the recomputation of an activation can influence the contributions of other activations, rendering a linear model insufficient. To validate this claim, we apply partial activation checkpointing to a ResNet18 model on a base Edge TPU accelerator, as detailed in Section~\ref{pag:resnet18}.

In this experiment, we evaluate four scenarios: recomputing no activations, recomputing the first and second activations that are used during the backpropagation and that are generated by the first layers $\text{AC}10$ and $\text{AC}01$, and both activations $\text{AC}11$. We maintain a consistent mapping configuration across all experiments and employ the layer-fusion algorithm proposed in Section~\ref{pag:lfs} to maximize fusion potential and data reuse. The results, adjusted by subtracting the baseline scenario of recomputing no activations, are presented in Figure~\ref{fig:resnetnonlinearity}.

Our findings reveal that the activation checkpointing problem cannot be accurately modeled using a linear MILP approach. Specifically, the recomputation cost of two activations is not equivalent to the sum of their individual recomputations. This non-linearity can stem from an improved data locality when recomputing activations immediately before gradient computation or a change in the layers that are fused together. Additionally, recomputation reduces the number of subgraphs with multiple output nodes, thereby increasing the number of layers that can be fused, as visualized in Figure~\ref{fig:activationCheckpointing}, where Op2 and Op3 can be fused together after applying activation checkpointing, which was not the case before.

A naive extension of the MILP formulation to accommodate fused-layer networks would require evaluating all possible combinations of activation orders, resulting in a combinatorial complexity of $\Theta(n \cdot n!)$, where $n$ is the size of the activation set $\mathcal{A}$. This complexity is prohibitive for both coefficient computation and solving the MILP problem itself. Alternatively, the problem could be modeled as a Mixed Integer Non-Linear Programming (MINLP) formulation but would be impractical to formulate and to solve. 

\subsubsection{Proposed Solution}
To address these challenges, we propose to use a genetic algorithm-based approach to generate a Pareto-optimal set of solutions, balancing memory reduction, energy consumption, and latency. This method avoids the intractable complexity of exhaustive MILP formulations while providing a practical framework for optimizing fused-layer neural networks. The energy and latency metrics are derived from Stream, and the memory metric is calculated as the sum of the activations' memory footprint, assuming FP16 storage. The genetic algorithm is based on NSGA-II~\cite{NSGAII2002}, a widely used multi-objective optimization algorithm that leverages elitism to preserve high-quality solutions across generations, while its crowding distance mechanism ensures diversity and a well-distributed Pareto front. This algorithm is also used in Stream for the scheduling optimization problem and makes it a good candidate to reuse.

Figure~\ref{fig:GA_AC} presents the results of this exploration for training a ResNet-18 model with the ADAM optimizer, using a batch size of 1 and an image size of 224$\times$224. The genetic algorithm successfully generates a Pareto front that balances energy, latency, and memory requirements. Notably, for an additional cost of 4\% in latency and energy, up to 13 MBs of memory can be saved. This effect becomes more pronounced with larger batch sizes, image sizes, or sequence lengths in language models. Furthermore, specific configurations yield lower memory and latency compared to the baseline, though not always lower energy costs. These results are consistent with activation checkpointing techniques used in PyTorch~\cite{pytorch_mincut_2022}.
% Further refinement of the genetic algorithm parameters could yield even better results.

This approach highlights the potential of activation checkpointing, when combined with layer fusion, to reduce the training cost of complex neural network architectures.

\section{Conclusion}\label{sec:conclusion}

This paper introduces MONET, which builds upon the Stream framework to enable modeling and optimization of full neural network training workloads on heterogeneous dataflow accelerators. By incorporating forward and backward execution, optimizer steps, layer fusion, and activation checkpointing into a unified pipeline, MONET provides accurate latency, energy, and memory estimates for training, exposing trade-offs that are not visible in inference-only analyses. Our experiments with ResNet-18 and GPT-2 demonstrate that architectures optimized for inference are not necessarily efficient for training, underscoring the need for dedicated training-aware co-design.

We further proposed optimization strategies for two key challenges in training: fused-layer partitioning and activation checkpointing. For fused-layer optimization, we introduced a constraint-based fusion solver that enables effective fusion on large training graphs and, despite using a heuristic objective, consistently discovers better solutions than manually designed fusion configurations. We also showed that the activation checkpointing problem is inherently non-linear and cannot be accurately captured by linear formulations. To address this, we employed a genetic algorithm that captures the complex trade-offs between latency, energy, and memory, leading to improved Pareto-optimal solutions. Together, these techniques reveal new operating points and highlight the importance of jointly exploring hardware, mapping, and algorithmic choices. MONET establishes a foundation for systematic training-centric accelerator design and supports future research toward more efficient deep learning systems.

Future work includes validating and extending this pipeline on modern accelerator architectures such as GPUs—similarly to inference-oriented efforts like LLMCompass~\cite{LLMCompass}—as well as exploring dedicated hardware designs tailored to this approach. We also plan to develop more advanced optimization algorithms to improve mapping strategies for large-scale training workloads.

%%

\section*{Acknowledgments}
This research was funded, in whole or in part, by the French National Research Agency (ANR) under the project ANR-22-CE25-0006 and was performed using AI resources from GENCI-IDRIS.

% {\appendix[]
% }

\bibliographystyle{IEEEtran}
\bibliography{bibliography}

\newpage

\section{Biography Section}
If you have an EPS/PDF photo (graphicx package needed), extra braces are
 needed around the contents of the optional argument to biography to prevent
 the LaTeX parser from getting confused when it sees the complicated
 $\backslash${\tt{includegraphics}} command within an optional argument. (You can create
 your own custom macro containing the $\backslash${\tt{includegraphics}} command to make things
 simpler here.)
 

% \end{IEEEbiography}



\vfill

\end{document}